\documentclass[accepted]{uai2023} 

\usepackage[american]{babel}

\usepackage{natbib} 
\makeatletter
\newcommand{\printfnsymbol}[1]{%
  \textsuperscript{\@fnsymbol{#1}}%
}
\makeatother
\usepackage[T1]{fontenc}
\newtheorem{proposition}{Proposition}
\usepackage{graphicx}
\usepackage{subcaption}

\usepackage{mathtools} 
\usepackage{booktabs} 
\usepackage{tikz} 

\usepackage{amssymb} 

\newcommand{\X}{\mathcal{X}} 
\newcommand{\x}{\mathbf{x}} 
\newcommand{\z}{\mathbf{z}} 
\newcommand{\J}{\mathcal{J}} 
\newcommand{\C}{\mathcal{C}} 
\renewcommand{\z}{\mathbf{z}} 
\newcommand{\Q}{\mathbf{Q}} 
\newcommand{\R}{\mathcal{R}} 
\DeclarePairedDelimiter\set{\lbrace}{\rbrace} 
\DeclarePairedDelimiter\norm{\lVert}{\rVert} 
\DeclareMathOperator*{\argmax}{argmax} 
\DeclareMathOperator*{\Bin}{Bin} 
\DeclareMathOperator*{\Bern}{Bern} 
\renewcommand{\P}{\mathbb{P}} 

\usepackage[american]{babel}
\usepackage{xr} 

\usepackage{algorithm}
\usepackage{algorithmic}
\usepackage{bm}
\usepackage{float}

\title{Structure-Aware Robustness Certificates for Graph Classification}

\author[1]{Pierre Osselin\thanks{Equal contribution.}}
\author[1]{Henry Kenlay\printfnsymbol{1}}
\author[1]{Xiaowen Dong}
\affil[1]{%
    Department of Engineering Science, University of Oxford, Oxford, UK
}
  
\begin{document}
\maketitle
\begin{abstract}
Certifying the robustness of a graph-based machine learning model poses a critical challenge for safety. Current robustness certificates for graph classifiers guarantee output invariance with respect to the total number of node pair flips (edge addition or edge deletion), which amounts to an $l_{0}$ ball centred on the adjacency matrix. Although theoretically attractive, this type of isotropic structural noise can be too restrictive in practical scenarios where some node pairs are more critical than others in determining the classifier's output. The certificate, in this case, gives a pessimistic depiction of the robustness of the graph model. To tackle this issue, we develop a randomised smoothing method based on adding an anisotropic noise distribution to the input graph structure. We show that our process generates structural-aware certificates for our classifiers, whereby the magnitude of robustness certificates can vary across different pre-defined structures of the graph. We demonstrate the benefits of these certificates in both synthetic and real-world experiments.
\end{abstract}

\section{Introduction}

\begin{figure*}
\includegraphics[width=0.6\textwidth]{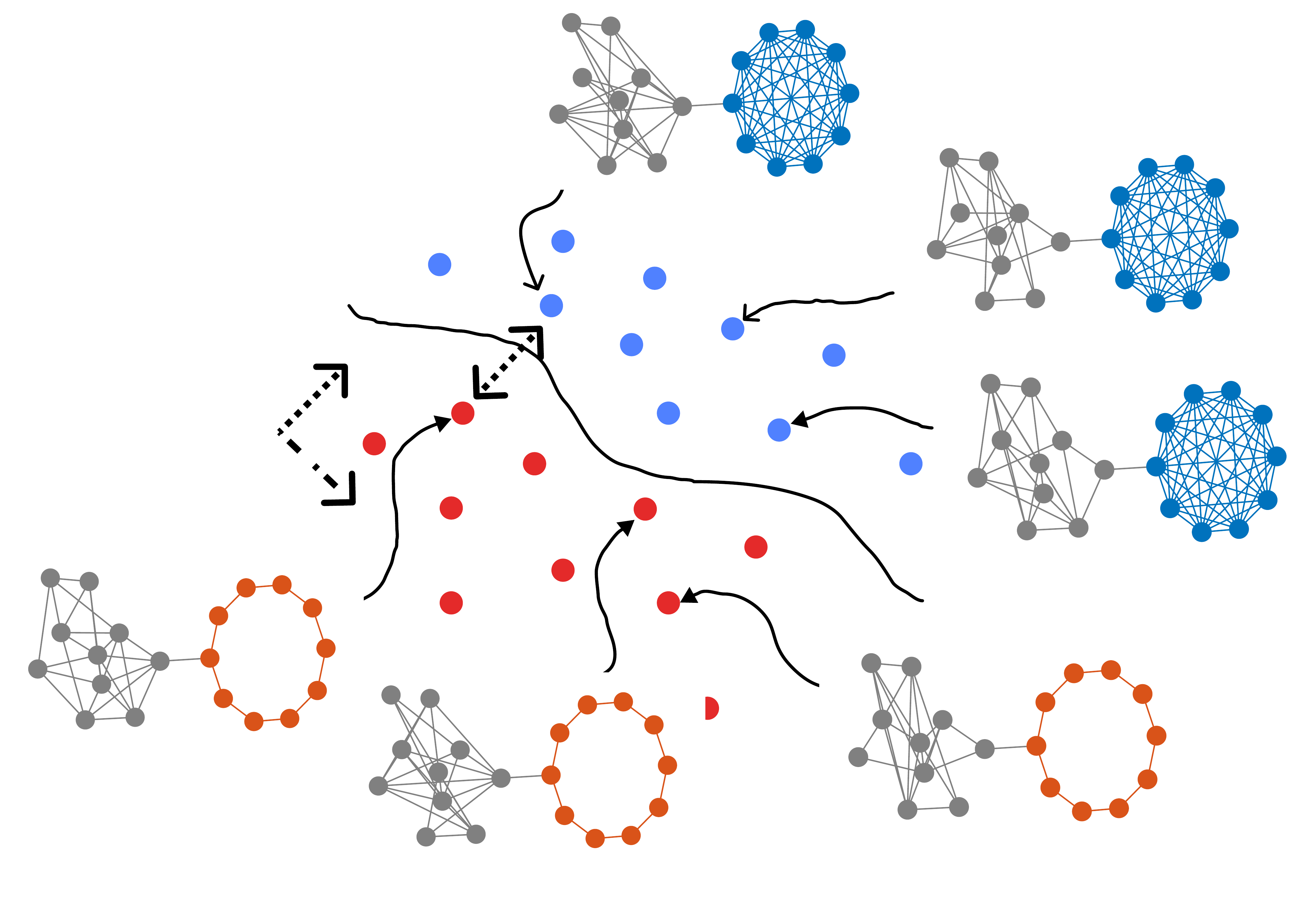}
\centering
\caption{An example of graph classification where the class label is determined by the blue or red part of the input graph, but not the grey part. In this case, perturbations to the former will be more likely to affect classification outcome than the latter. In other words, the classification decision boundary is sensitive to some perturbations to the input graph structure (which move the point towards the boundary), but robust to others (which keep the point away from the boundary).}
\label{fig:anisotropic}
\end{figure*}

Graph-based machine learning models have made considerable strides in the last couple of years, with applications ranging from NLP \citep{wu2023graph}, combinatorial optimisation \citep{drori2020learning} and protein function prediction \citep{gligorijevic2021structure}. As these tools become more common, studying their vulnerability to potential adversarial examples becomes paramount for safety.

Robustness certification is an active field of research whose goal is to develop certificates guaranteeing invariance of the model prediction with respect to some input perturbations. Diverse methods have been used to achieve this goal, from interval bound propagation \citep{gowal2019scalable}, convex relaxation \citep{raghunathan2018semidefinite}, Lipschitz bounds computation \citep{huang2021training} or randomised smoothing \citep{wang2021certified}. Given a data point $\x$ and a set of perturbed inputs $\mathcal{B}(\x)$, a robustness certificate verifies that a model's prediction $f(\x)$ remains unchanged for all other inputs in the perturbation set. That is, for all $\x' \in \mathcal{B}(\x)$ it holds that $f(\x)=f(\x')$. Often the  set of perturbed inputs $\mathcal{B}(\x)$ is parameterised, for example by a closed-ball $\mathcal{B}_{r}(\x) = \set{\x' : d(\x, \x') \leq r}$ under some distance function $d$ and radius $r$. In this case, we are interested in knowing the largest $r$ that we can certify for, where $r$ is called the certified radius. 


In the context of robustness certification of graph classifiers against structural perturbation, a common choice of perturbation set is the set of all graphs reachable from an input graph $\x$ by up to $r$ node pair flips (edge additions and deletions)\footnote{We use the terminology of node pair flip instead of edge flip to emphasise that we are considering the addition of edges that do not exist in the original graph as well as the deletion of existing edges.}. This corresponds to a closed ball on the upper triangle entries of the adjacency matrix where the distance is induced by the $\ell_1$ norm and the bottom triangle entries are determined by the constraint that the adjacency matrix is symmetric (assuming for simplicity the graph is unweighted and undirected). In some cases, however, different node pairs of the graph can be more predictive of the ground truth label than others. A real-world example is classification of molecular structures, where the edges that constitute key substructure (e.g., a ring) are more critical in determining the class label than the rest. A synthetic example is further presented in Fig.~\ref{fig:anisotropic}.
In such situations, certifying according to a total number of edge additions or deletions
might gives a pessimistic certified radius, because the set of perturbed inputs may include perturbations which consist of flipping many critical node pairs (in terms of determining the graph label).


In this work, we address this problem by defining disjoint regions of node pairs and proposing robustness certificates that verify that the prediction of the classifier will not change for a potentially different number of node pair flips for each region. Such disjoint regions may be either obtained via domain knowledge (as in the molecule example mentioned above) or assigned based on the level of importance of node pairs to the classification task.
Our approach then relies on randomised smoothing, which is a powerful framework to produce robustness certificates which hold with high probability. Given some noise distribution over the input, randomised smoothing transforms a base model $f$ into a smoothed model $g$ for which we can provide probabilistic robustness guarantees. 

Existing  randomised smoothing approaches for certifying graph classification mostly consider an isotropic noise distribution that flips each node pair with a fixed probability \citep{jia2020certified,gao2020certified,wang2021certified}. 
The certificate in this case corresponds to the total number of node pair flips. Instead, we propose using an anisotropic noise distribution based on the predefined regions whereby the probability of flipping a node pair depends on to which region (if any) the node pair belongs. We show that smoothed classifiers constructed using this anisotropic noise distribution naturally lead to structure-aware robustness certificates whereby different numbers of node flips are certified for each of the regions. We demonstrate \footnote{Code to reproduce our numerical experiments can be found at \href{https://github.com/pierreosselin/structureaware.git}{https://github.com/pierreosselin/structureaware.git}} the benefits of our approach on both synthetic and real-world experiments.

To the best of our knowledge, our method is one of the first of its kind that allows for flexible and structure-aware graph certification in the input graph domain.

\section{Preliminaries}

\begin{figure}
\includegraphics[width=0.4\textwidth]{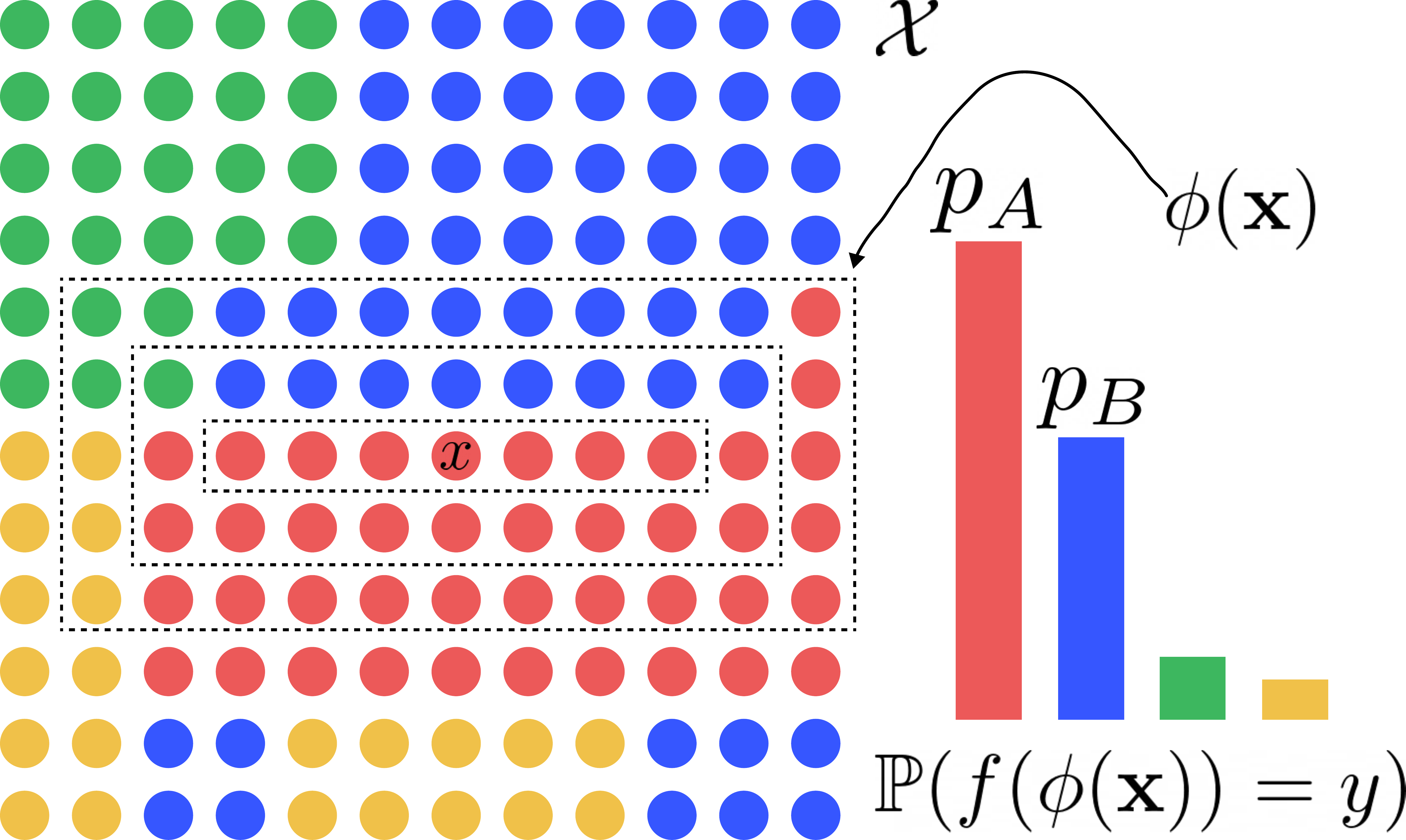}
\centering
\caption{Illustration of a smoothed classifier: at every point $\x$ a neighborhood vote is performed according to a distribution $\phi(\x)$ centered on $\x$. The figure was inspired by \citet{gunnemann20}.}
\label{fig:smoothed_classifier}
\end{figure}

Let $\mathcal{X}$ be a data space and $f : \mathcal{X} \rightarrow \mathcal{Y}$ be a classifier which maps each point $\x \in \X$ to a label $y \in \mathcal{Y}$. Let $\phi : \mathcal{X} \rightarrow \mathcal{P(X)} $ be a noise distribution over our data, that is, $\phi(\x)$ returns a distribution over $\mathcal{X}$ for every point $\x \in \X$.
We write $f(\phi(\x))$ to denote the random variable $\mathbb{P}_{\z \sim \phi(\x)}(f(\z))$. This represents the distribution of outputs of the base classifier given the randomisation scheme applied to the input. We define $g$ to be the smoothed classifier of our base classifier $f$ as
\begin{equation}
\label{eq:smoothclassifier}
    g(\x) = \argmax_{y \in \mathcal{Y}} \mathbb{P}(f(\phi(\x)) = y).
\end{equation}
The smoothed classifier can be interpreted as a neighbourhood vote, where the output is the mode of the output of the classifier when inputs are sampled from the distribution $\phi(\x)$. An illustration of a smoothed classifier is given in Fig.~\ref{fig:smoothed_classifier}.

\subsection{Certifying a smoothed classifier}
We can construct a lower and upper bound $\underline{\rho_{\x, \tilde{\x}}}(p,y)$ and $\overline{\rho_{\x, \tilde{\x}}}(p,y)$ on $\mathbb{P}(f(\phi(\tilde \x))=y)$, which is the probability of class $y$ under our classifier $f$ smoothed by $\phi$ and evaluated on an arbitrary point $\tilde \x \in \mathcal{X}$:
\begin{equation}
    \underline{\rho_{\x, \tilde{\x}}}(p,y) = \min\limits_{\substack{h \in \mathcal{H}: \\ \mathbb{P}(h(\phi(\x)) = y) = p}} \mathbb{P}(h(\phi(\tilde{\x})) = y)
    \label{eq:lee_lower}
\end{equation}
\begin{equation}
    \overline{\rho_{\x, \tilde{\x}}}(p,y) = \max\limits_{\substack{h \in \mathcal{H}: \\ \mathbb{P}(h(\phi(\x)) = y) = p}} \mathbb{P}(h(\phi(\tilde{\x})) = y)
    \label{eq:lee_upper}
\end{equation}
In this definition, $\mathcal{H}$ is the set of measurable classifiers with respect to $\phi$. Because the optimisation constraints include the base classifier $f \in \mathcal{H}$ it follows that 
\begin{equation}
    \underline{\rho_{\x, \tilde{\x}}}(p,y) \leq \mathbb{P}(f(\phi(\tilde \x))=y) \leq \overline{\rho_{\x, \tilde{\x}}}(p,y).
\end{equation}


We define a perturbation set $\mathcal{B}_r(\x)$ as a family of sets $\mathcal{B}_r(\x) \subseteq \X$ parameterised by some $r \geq 0$ such that  $\mathcal{B}_r(\x) \subseteq \mathcal{B}_{r'}(\x)$ if and only if $r \leq r'$. This includes open or closed balls with respect to a metric over $\X$. We say that the smoothed classifier $g$ is certified at $\x$ in some perturbation set $\mathcal{B}_r(\x)$ if the output of $g$ is the same for all neighbouring points $\tilde \x \in \mathcal{B}_r(\x)$ in the perturbation set. 

Following \citet{cohen2019certified}, we will write $c_A$ to be the output class of $g(\x)$ which is returned with probability $p_A$, and $c_B$ to be the ``runner-up'' class, i.e., the class distinct from $c_A$ with the next highest probability $p_B$. From the framework of \citet{lee2019tight}, we define the notion of a point-wise certificate around a point $\x$ by verifying if the following holds:
\begin{equation}
    \label{eq:margin_all}
    \min\limits_{\tilde{\x} \in \mathcal{B}_{r}(\x)} \Phi_{\x, \tilde{\x}} (p_A, c_A) > 0,
\end{equation}
where
\begin{equation}
    \label{eq:margin}
    \Phi_{\x, \tilde{\x}} (p_A, c_A) \triangleq 
    \underline{\rho_{\x, \tilde{\x}}}(p_A, c_A) - \overline{\rho_{\x, \tilde{\x}}}(p_B, c_B).
\end{equation}
Eq.~(\ref{eq:margin}) can be thought of as a margin, which gives the difference between a lower bound on $p_A$ and an upper bound on $p_B$ for some point $\tilde \x$ in the perturbation set. Eq.~(\ref{eq:margin_all}) then indicates if this property holds for all $\tilde \x$ in the perturbation set.

\subsection{Certified radius}
We can define the certified radius to be the largest value of $r$ so that we can certify with respect to the predefined perturbation set $\mathcal{B}_r(\x)$. Formally this can be defined as:
\begin{equation} \label{opti2}
    R(\x) = \sup r \text{, s.t. } \min\limits_{\tilde{\x} \in \mathcal{B}_{r}(\x)} \Phi_{\x, \tilde{\x}} (p_A,c_A) > 0.
\end{equation}

If the perturbation set is an open or closed ball, $R(\x)$ is the radius of the largest ball we can certify over. 

\subsection{Computing the certificate}
\subsubsection{Computing a point-wise certificate}
For a fixed $\x$ and neighbouring point $\tilde{\x}$, we can compute Eq.~(\ref{opti2}) following the method described by \citet{lee2019tight}.  First, we partition the space $\mathcal{X} = \bigcup_{i} \mathcal{R}_{i}$ into disjoint regions of equal likelihood ratios $\mathcal{R}_{k} = \{\z \in \mathcal{X}: \frac{\P(\phi(\tilde{\x}) = \z)}{\P(\phi(\x) = \z)} = \eta_{k} \}$ where without loss of generality we can assume $\eta_{k} \in \mathbb{R}$ are in an ascending order. The quantity $\Phi_{\x, \tilde{\x}} (p_A,c_A)$ can then be computed by solving the following linear programming (LP) problems \citep[Lemma 2]{lee2019tight}:    
\begin{equation}
    \label{eq:soln_upper}
    \underline{\rho_{\x, \tilde{\x}}}(p_A,c_A) = \min_{\mathbf{h}} \mathbf{h}^T \mathbf{\tilde{r}} \quad \text{ s.t. } \quad  \mathbf{h}^{T}\mathbf{r} = p_A, \quad  0 \leq \mathbf{h} \leq 1
\end{equation}
and similarly,
\begin{equation}
    \label{eq:soln_lower}
    \overline{\rho_{\x, \tilde{\x}}}(p_B, c_B) = \min_{\mathbf{h}} \mathbf{h}^T \mathbf{\tilde{r}} \quad \text{ s.t. } \quad  \mathbf{h}^{T}\mathbf{r} = p_B, \quad  0 \leq \mathbf{h} \leq 1.
\end{equation}

The variable $\mathbf{h}$ corresponds to optimising over the classifiers that are optimised over in Eq.~(\ref{eq:lee_lower}) and Eq.~(\ref{eq:lee_upper}). The vectors $\mathbf{r}$ and $\tilde{\mathbf{r}}$ are $\mathbf{r}_{i} = \mathbb{P}(\phi(\x) \in \mathcal{R}_{i})$ and $\tilde{\mathbf{r}}_{i} = \mathbb{P}(\phi(\tilde{\x}) \in \mathcal{R}_{i})$ respectively. This LP problem can be solved via a greedy approach \citep{lee2019tight}. Given the ratios $\eta_{k}$ are ordered in an ascending manner, starting with $\mathbf{h}=\mathbf{0}$ we can assign $\mathbf{h}_i=1$ for the indices $i=1,\ldots, k-1$ such that we choose the largest $k$ with $\mathbf{h}^{T}\mathbf{r} \leq p_A$, and set the value of $\mathbf{h}_{k}$ such that $\mathbf{h}^{T}\mathbf{r}=p_A$. We can solve Eq.~(\ref{eq:soln_upper}) and Eq.~(\ref{eq:soln_lower}) in a similar way. 

Given this efficient way to compute a certificate, there remain some quantities that must be calculated. For our certificate, we introduce the methods to compute them in Section \ref{sec:framework}. The first is partitioning the space $\X$ into disjoint unions of equal likelihood ratios. We provide a method to do so in Proposition \ref{prop:disjoint_unions}. Next, the values $\eta_k$ must be computed, or at least given in closed form, so the regions can be sorted in ascending order. Proposition \ref{prop:ratio} gives an analytic closed-form. Finally, a closed form for $\P(\phi(\x) = \z)$ allows us to compute $\mathbf{r}$. We can compute  $\mathbf{\tilde{r}}$, required to solve the linear programs, by noticing that $\P(\phi(\tilde{\x}) = \z) = \eta_k \P(\phi(\x) = \z)$. We compute $\P(\phi(\x) = \z)$ in Proposition \ref{prop:region}.
\subsubsection{Computing a regional certificate}
To certify over some allowed set $\mathcal{B}_r(\x)$ we need to find the worst case $\tilde{\x} \in \mathcal{B}_r(\x)$ to solve Eq.~(\ref{eq:margin_all}). Through particular choices of parameterising $\tilde{\x}$, one can find equal likelihood regions that give rise to equal likelihood ratios values that are independent of the exact value of $\tilde{\x}$, turning the point-wise certificate into a regional certificate, where the region is given by specific parameterisation. We give such an example in Proposition \ref{prop:ratio}. 

\subsection{Randomised smoothing for graph classification} 
Although our method is applicable to any setting with discrete data, we are interested in the robustness of graph classification models to structural perturbation of undirected, unweighted graphs\footnote{This work can be extended to the setting of directed graphs as well as the task of node classification.}.
In this case, the input domain is the set of finite graphs $\mathcal{X} = \cup_{i=1}^{\infty} \X_n$ where $\X_n$ is the space of graphs with $n$ nodes. We can represent a graph with $n$ nodes as a binary vector of size $n \choose 2$ where each entry indicates the presence or absence of an edge. Without loss of generality we will treat graphs as binary vectors $\mathbf{x}$ from here on in\footnote{Note that due to isomorphism multiple different binary vectors can represent the same graph.}.  


We are interested in robustness with respect to node pair flips which can represent an edge addition (change a zero to a one in $\mathbf{x}$) or edge deletion (change a one to a zero). This may be interpreted as adding ``structural noise'' to the input graph. Two candidates for the distribution of such noise have been proposed in the literature.
The first one applies an independent Bernoulli distribution to every node pair. That is, $\phi(\x)_{i} = \x_{i} \oplus \mathbf{\epsilon}_{i}$ with $\mathbf{\epsilon}_{i} \sim \Bern(p)$ for a fixed $p$, where $\oplus$ is the bitwise XOR operator. We refer to this noise distribution as isotropic, as it flips each node pair with equal probability. This approach is used by \citet{jia2020certified,wang2021certified,gao2020certified}. 
The second approach, a sparsity-aware noise distribution \citep{bojchevski2020efficient}, gives a different probability of edge flipping depending on whether the edge is present in the graph. If an edge exists between a node pair then it is flipped with probability $p_-$, whereas if a node pair does not exist between two nodes it is added with probability $p_+$. This distribution can be written as  ${P(\phi(\x)_{i} \neq \x_{i}) = p_{-}^{\x_{i}}p_{+}^{1-\x_{i}}}$. The proposed framework is conceptually similar to this latter case; however,
we consider a generic partition of node pairs into one of many node pair sets and these pairs are perturbed according to their membership of the sets. This leads to the derivation of robustness certificates that are aware of the structural characteristics of the input graph with respect to the classification labels.

\section{Randomised smoothing with anisotropic noise}
\label{sec:framework}
Given $\x \in \mathcal{X}_{n}$, suppose we divide our input space of node pairs up into disjoint regions $\bigsqcup_{i \in I} \C_{i}$ such that each node pair belongs to exactly one region and there are a total of $C$ regions. We define a noise distribution where independent Bernoulli distributions are applied to every node pair, and where the parameter of the Bernoulli distribution is shared within every set $\C_{i}$:
\begin{equation}
\phi(\x)_{k}  = \x_{k} \oplus \mathbf{\epsilon}_{k} \text{, where } \mathbf{\epsilon}_{k} \sim \Bern(p_{i}) \text{ and } k \in \C_{i}, 
\end{equation}

Let $\mathbf{R} \in \mathbb{Z}^C$ be a tuple of integers such that $0 \leq \mathbf{R}_i \leq |\C_i|$ and let $\mathcal{B}_{\mathbf{R}}(x) = \set{\z \in \X_{n} : \norm{\z_{\C_{i}} - \x_{\C_{i}}}_0 \leq \mathbf{R}_{i} }$ be a perturbation set. Let $\tilde \x \in \mathcal{B}_{\mathbf{R}}(x)$ and $\J = \{ i : \x_i \not = \tilde \x_i \}$ be indices of $\x$ which are perturbed to give $\tilde \x$. Furthermore, let $\J_i = \J \cap \C_{i}$ be indices where $\x$ is perturbed in collection $\C_i$. In our case a maximum radius means maximising the radius on every regions $\C_i$.

First, we develop a set decomposition on which likelihood ratios can be computed.
\begin{proposition}
\label{prop:disjoint_unions}
We define regions $\R_\Q = \set{\z \in \X_{n} : \norm{\z_{\J_{i}} - \x_{\J_{i}}}_0 = Q_{i} : i \in I}$ that represent points $\z$ which agree with $\x$ by exactly $Q_{i}$ bits in sub-regions $\J_{i}$. Then $\X_{n}$ can be represented by the following disjoint union
\begin{equation}
    \bigcup_{\mathbf{0} \leq \mathbf{Q} \leq \mathbf{R}} \R_{\mathbf{Q}},
\end{equation}
where vector inequalities are element-wise.
\end{proposition}

Next, the likelihood ratio is fixed for elements $\z$ in any one region. This likelihood ratio has the following closed form.
\begin{proposition}
\label{prop:ratio}
Consider a region $\R_\Q = \set{\z \in \X_{n} : \norm{\z_{\J_{i}} - \x_{\J_{i}}}_0 = Q_{i}}$ then for all $\z \in \mathcal{R}_{\Q}$  the following holds 
\begin{equation}
    \eta^{\mathcal{R}}_\mathcal{\Q}= \frac{\P(\phi(\tilde{\x}) = \z)}{\P(\phi(\x) = \z)} = \prod_{i=1}^C \bigg( \frac{1 - p_{i}}{p_{i}} \bigg)^{R_{i} - 2Q_{i}}.
\end{equation}
\end{proposition}
Finally, we can compute the likelihood of a smoothed input belonging to these regions:
\begin{proposition}
\label{prop:region}
The probability of the output of a smoothed input $\phi(\x)$ belonging to a region $\R_\Q$ is given by
\begin{equation}
    \P(\phi(\x) \in \mathcal{R}_{\Q}) = \prod_{i = 1 }^C \Bin(R_i - Q_{i} | R_{i}, p_{i}),
\end{equation}
where $\Bin(R_i -Q_{i} | R_{i}, p_{i})$ is the probability mass function of the binomial distribution giving probability of $R_i - Q_{i}$ successes from $ R_{i}$ trials each with success probability $p_{i}$.
\end{proposition}
All proofs are provided in Appendix 1.
Using these results we can provide robustness certificates of the smoothed classifier. Given $\x \in \X$ and our noise distribution, we can compute the values $p_A$ and $p_B$. In practice, these quantities are not available in closed form and are estimated via sampling, as in \citet{bojchevski2020efficient}. A more detailed description is given in Appendix 2, which gives probabilistic certificates according to some confidence intervals. Without loss of generality we order the regions as $\R_1, \ldots \R_T$ (where $T = \prod_i (R_i + 1))$. The corresponding ratios $\eta_{\Q}^{\R}$ as given by Proposition \ref{prop:ratio} are ordered as $\eta_1 \leq \ldots \leq \eta_T$. From these elements, $\Phi_{\x, \tilde{\x}} (p_A,c_A)$ can be computed through Eq.~(\ref{eq:margin}) and the optimisation problem Eq.~(\ref{opti2}) can be solved. In practice, the optimisation of Eq.~(\ref{opti2}) can be solved efficiently by leveraging symmetries displayed by $\Phi_{\x, \tilde{\x}} (p_A,c_A)$ when $\tilde{\x}$ varies. This property is more thoroughly described in Appendix 2. A complete description of the algorithm to compute the proposed certificate is provided in Appendix 3. The computational complexities of verifying Eq.~(\ref{eq:margin_all}) and solving Eq.~(\ref{opti2}) are also provided in Appendix 3.

\section{Related work}
\textbf{Robustness certificate.}
The earliest work proposing a robustness certificate for graph-based models is \citet{zugner2019certifiable}, where the authors consider semi-supervised node classification over graphs with binary attributes using graph neural networks. The admissible perturbations are those bounded under the $\ell_0$ semi-norm. Following the approach of \citet{wong2018provable}, a convex relaxation is considered and a dual problem  is solved via linear programming. 
Other early works that consider certificates under topological change include \citet{bojchevski2019certifiable} who propose a certificate for a class of linear graph neural networks based on PageRank diffusion. \citet{jin2020certified} propose the first certificate for graph classifiers which consist of a single graph convolutional layer followed by a pooling layer and a final linear layer. A convex relaxation of the adversarial polytope based on Lagrange duality is considered. Although the computed certificates are exact, the framework is specific to this simple graph classifier model and is expensive to compute. Recently, \citet{jin2022certifying} consider certifying graph classifiers under a different admissible perturbation measure, i.e., the orthogonal Gromov-Wasserstein discrepancy.

\textbf{Randomised smoothing based approaches.}
Randomised smoothing was originally proposed by \citet{lecuyer2019certified} and \citet{singla2020second} in the context of image classification. \citet{cohen2019certified} study this approach further, giving a tight certificate for perturbations bounded by an $\ell_2$ norm 
and improving the scalability of the approach.
They also outline some disadvantages of randomised smoothing, such as the need for high levels of noise or a substantial number of samples to certify to a large certified radius. Another consideration of randomised smoothing is their accuracy-robustness trade-off, controlled by the level of noise. 

Existing  randomised smoothing approaches for certifying graph-based classification models consider certificates for the number of edge flips. This is achieved by flipping an edge between each node pair independently with the same probability $p$. \citet{jia2020certified} apply this to community detection algorithms, whereas \citet{gao2020certified} and \citet{wang2021certified} apply the same principles to graph classification.
The closest method from our work is the sparsity-aware certificate proposed in \citet{bojchevski2020efficient}. The authors note that due to the sparse nature of many real-world graphs, adding and deleting edges between node pairs with the same probability leads to many more edges being added than deleted. To account for this, the authors propose using different probabilities for adding and deleting edges. This method is different from ours as the Bernoulli probabilities change with the input graph whereas ours depend on predefined sets of node pairs. These sets in our case account for varying levels of importance of the node pairs in predicting a graph label.


\section{Experiments}

\subsection{Synthetic experiment}
\begin{figure*}
\centering
  \begin{tabular}[b]{@{}c@{}}
    \begin{subfigure}[b]{0.35\textwidth}
      \includegraphics[width=\textwidth]{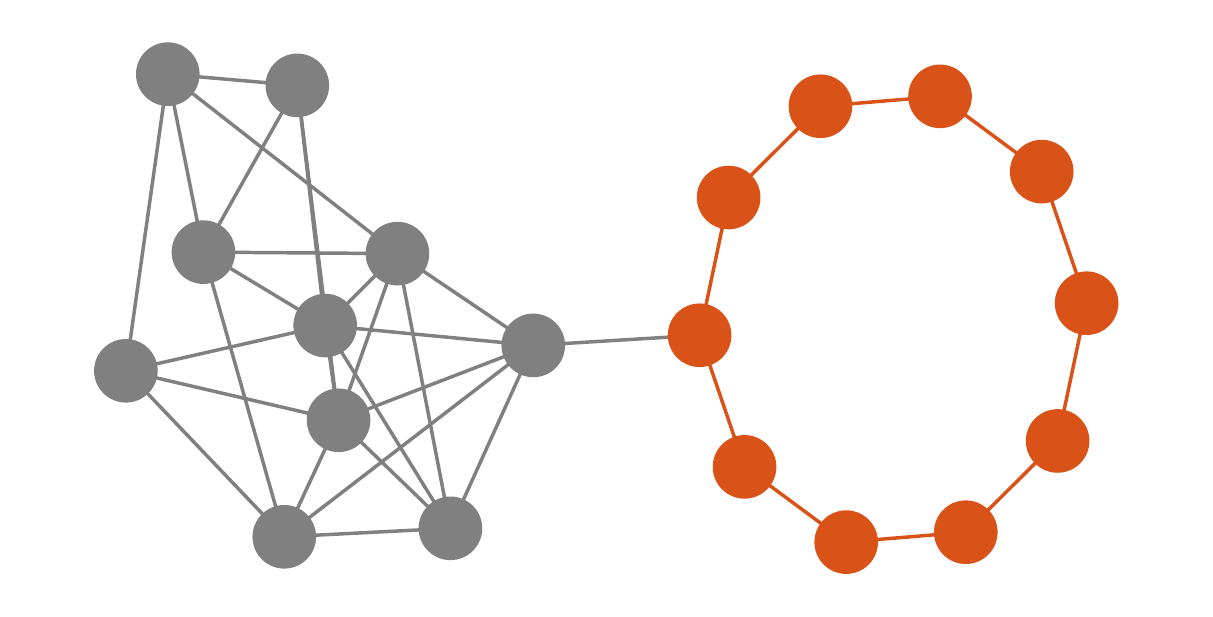}
      \caption{Graph with negative label.}\label{fig:graphs_and_noise_graph_0}
    \end{subfigure}\\
    \begin{subfigure}[b]{0.35\textwidth}
      \includegraphics[width=\textwidth]{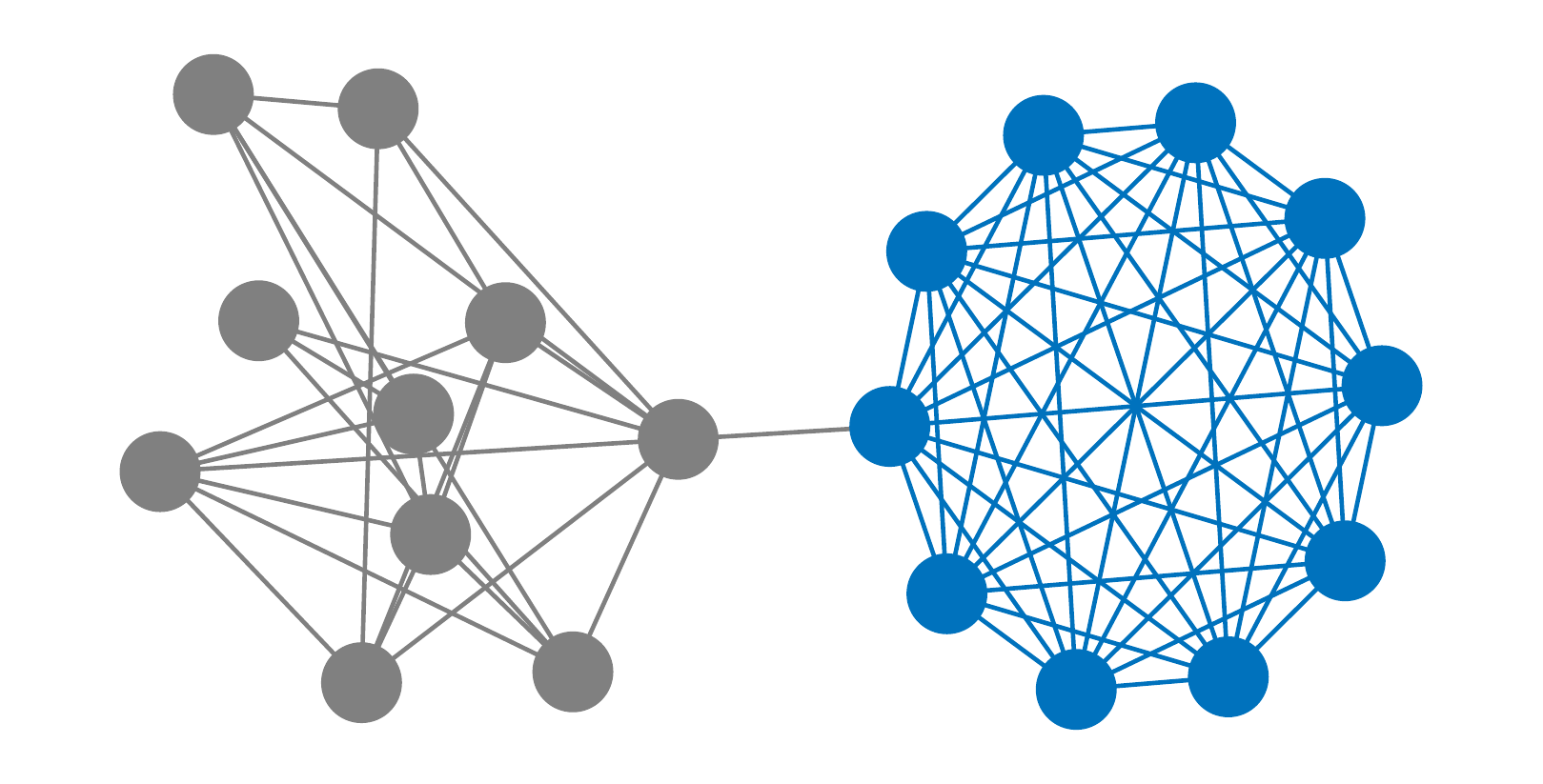}
      \caption{Graph with positive label.}\label{fig:graphs_and_noise_graph_1}
    \end{subfigure}
  \end{tabular}
  \begin{subfigure}[b]{0.4\textwidth}
    \includegraphics[width=\textwidth]{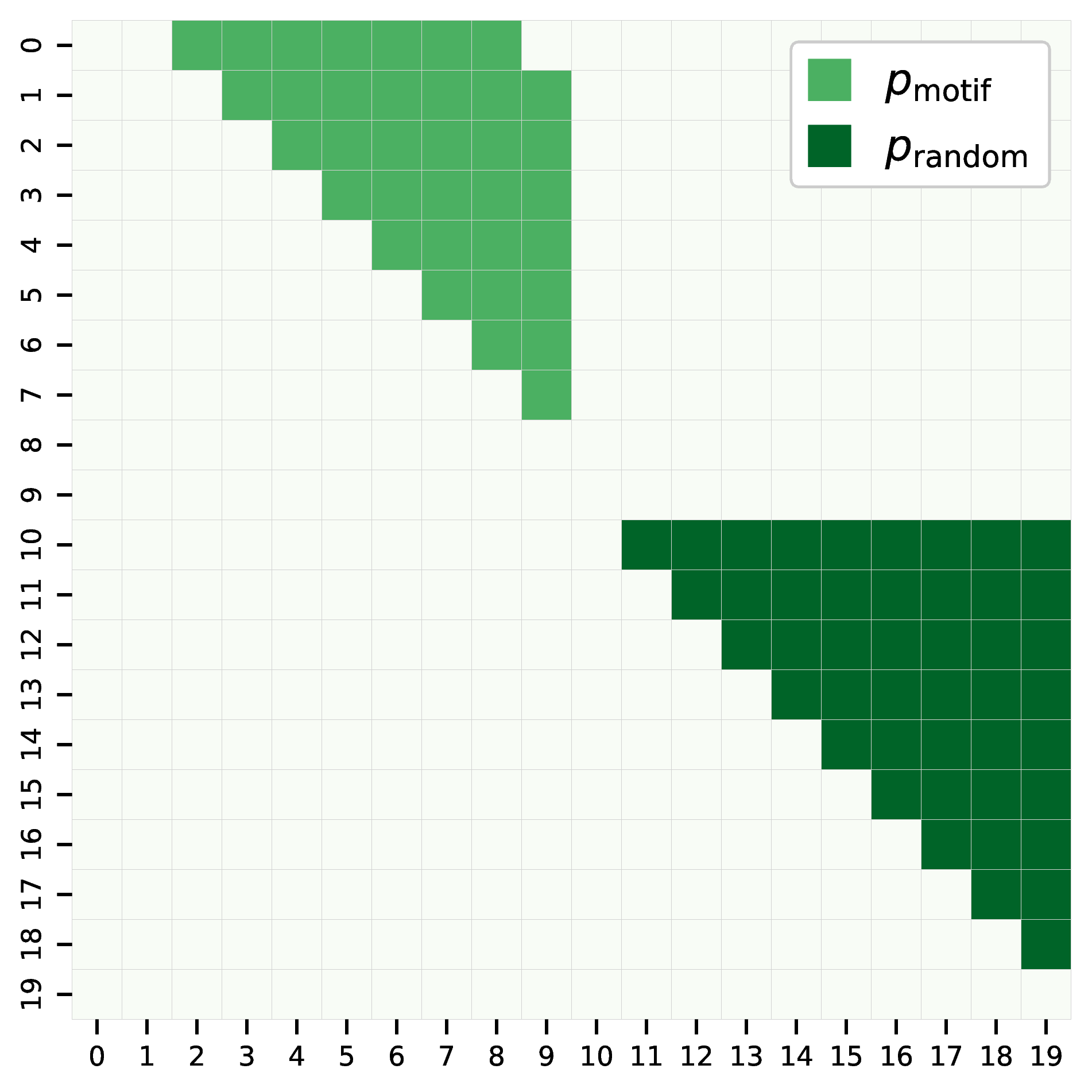}
    \caption{Noise matrix.}\label{fig:noise_matrix}
  \end{subfigure}
  \caption{Example graphs with a positive and negative label. Blue nodes and edges denote a motif part of a positive label and red nodes and edges denote a motif part of a negative label. The noise matrix show how edges are perturbed with $p_{\text{motif}}$ being the noise parameter for node pairs in the motif part and $p_{\text{random}}$ being the noise applied to node pairs in the random part. Notice that only the internal edges of the motif are perturbed, and the bridge edge is not perturbed. Only the upper triangle of the noise is shown only; in practice this is sampled and applied to the upper and lower triangle of the adjacency matrix so the graph remains undirected.}
  \label{fig:graphs_and_noise}
\end{figure*}
We motivate the use of anisotropic noise by considering inputs $\x$ that are an element of some space $\X = \X_1 \oplus \X_2$ where $\oplus$ is the direct sum. Consider a point that is close to the decision boundary in the $\X_1$ subspace but far from the decision boundary in the $\X_2$ subspace as illustrated in Fig.~\ref{fig:anisotropic}. An isotropic certificate cannot certify beyond the small distance to the decision boundary in $\X_1$. However, by design our certificate can certify the distances of $\X_1$ and $\X_2$ jointly allowing us to certify a small distance in the $\X_1$ subspace but a large distance in the $\X_2$ subspace. 

We design a synthetic graph classification data set whereby the graphs are constructed using a motif which determines the class label (corresponding to the important subspace $\X_1$) connected to a randomly generated graph (corresponding to the unimportant subspace $\X_2$) which is independent of the class label. We can consider edges in the motif part to be in one node pair set $\C_{\text{motif}}$ and edges from the random part in $\C_{\text{random}}$. Given a model that solves this task, we would expect changes in the motif to move the input closer or further away from a decision boundary but changes in the random part of the graph to move the point parallel to the direction of the decision boundary. In other words, we should be able to certify a large number of node pairs in $\C_{\text{random}}$ by applying a large value of noise $p_{\text{random}}$ without hurting the accuracy of the smoothed classifier. We cannot certify a large number of node pairs in $\C_{\text{motif}}$ without a drop in accuracy, so we choose a small value of noise $p_{\text{motif}}$ to retain high accuracy. 

More concretely, we generate a binary classification problem where each graph has a motif part of $n_{\text{motif}}=10$ nodes where a cycle determines a negative label and a complete graph determines a positive label. We then generate a random part using a connected Erd\H{o}s-R\'enyi graph \citep{gilbert1959random} with $n_{\text{random}}=10$ nodes and parameter $p=0.5$. We join these graphs using a single edge. See Fig.~\ref{fig:graphs_and_noise_graph_0} for an example of the negative class and Fig.~\ref{fig:graphs_and_noise_graph_1} for an example of the positive class. 

We generate balanced train, validation and test sets of size $1000$, $1000$, and $100$ respectively. The test set is smaller as the randomised smoothing procedure is computationally expensive. This is because to estimate $p_A$ a large number of random inputs need to be generated and classified using the model. 
We train as a base classifier an SVM using a node label histogram kernel \citep{sugiyama2015halting} where the node label corresponds to the node degree. Let $c(\mathcal{G}, d)$ be a function that counts the number of nodes in a graph $\mathcal{G}$ with degree $d$. Then the kernel applied to graphs $\mathcal{G}_1$ and $\mathcal{G}_2$ can be written as $\kappa(\mathcal{G}_1, \mathcal{G}_2) = \sum_{d=0}^\infty c(\mathcal{G}_1, d) \cdot c(\mathcal{G}_2, d)$ which is well defined for finite graphs. We use this model as we expect it to be sensitive to the motif structure that determines the label; the negative label gives a large value in the $c(\cdot, 2)$ component whereas the positive label gives a large value in the $c(\cdot, n_\text{motif}-1)$ component. Indeed, the base classifier gets $100\%$ accuracy on the train, validation and test data sets. We present further results for different choices of kernel in Appendix 4.

For the certification procedure, we apply noise separately for node pairs in the motif part and node pairs in the random part. We apply noise to the internal edges of the motif part only (and not to those of the red cycle, see Fig.~\ref{fig:graphs_and_noise}). We do not apply noise to node pairs where one node lies in the motif and one does not. The noise matrix is shown in Fig.~\ref{fig:noise_matrix}. We generate $100,000$ perturbations per test sample and use these to estimate the output of the smoothed classifier and generate a certificate. We experiment with varying the number of perturbations in Appendix 4. We use a confidence level of $\alpha=0.99$ to estimate $p_A$. We also compute certificates using isotropic noise for comparison. 

For our certificate with anisotropic noise we consider $\mathbf{p} = (p_{\text{motif}}, p_{\text{random}}) \in \mathbf{P}$ where $\mathbf{P}= \set{0.02, 0.04, \ldots, 0.2} \times \set{0.05, 0.1, \ldots, 0.45}$. Recall that $p_{\text{motif}}$ is the noise parameter for the motif part and $p_{\text{random}}$ is the noise parameter for the random part. For the isotropic certificate we consider $p \in \set{0.02, 0.04, \ldots, 0.2}$. For the anisotropic certificate, we certify over perturbation pairs $\mathbf{r}=(r_{\text{motif}}, r_\text{random})$ which means that with high probability $r_{\text{motif}}$ edge flips in the motif part and $r_{\text{random}}$ edge flips in the random part will not change the label of the smoothed classifier. This is different from the isotropic certificate which guarantees the label does not change for $r$ edge flips anywhere in the graph.

\begin{figure*}
\centering
    \begin{subfigure}[b]{0.45\textwidth}
    \includegraphics[width=\textwidth]{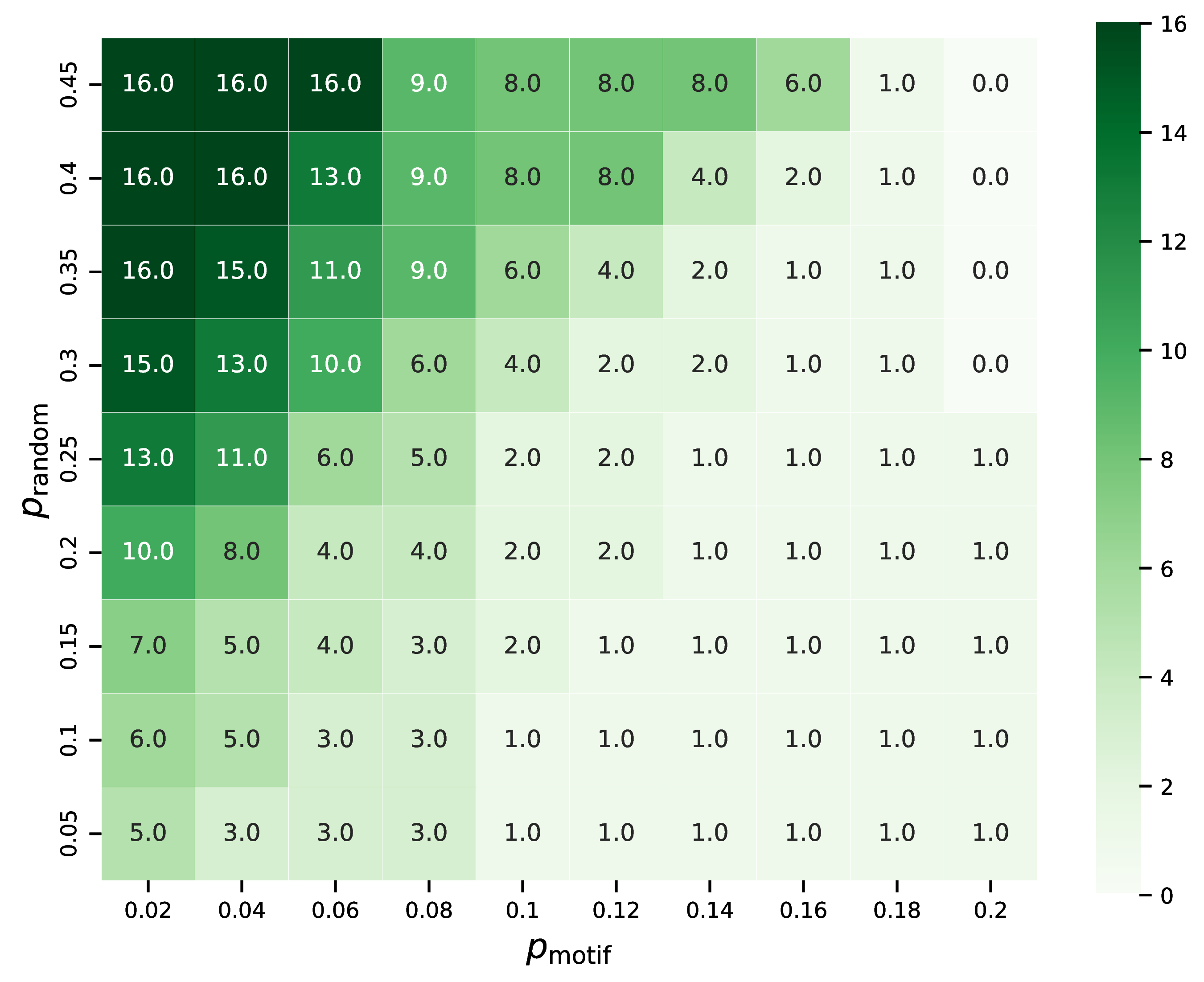}
    \caption{Score.}
    \label{fig:score_heatmap}
    \end{subfigure}
    \begin{subfigure}[b]{0.45\textwidth}
    \includegraphics[width=\textwidth]{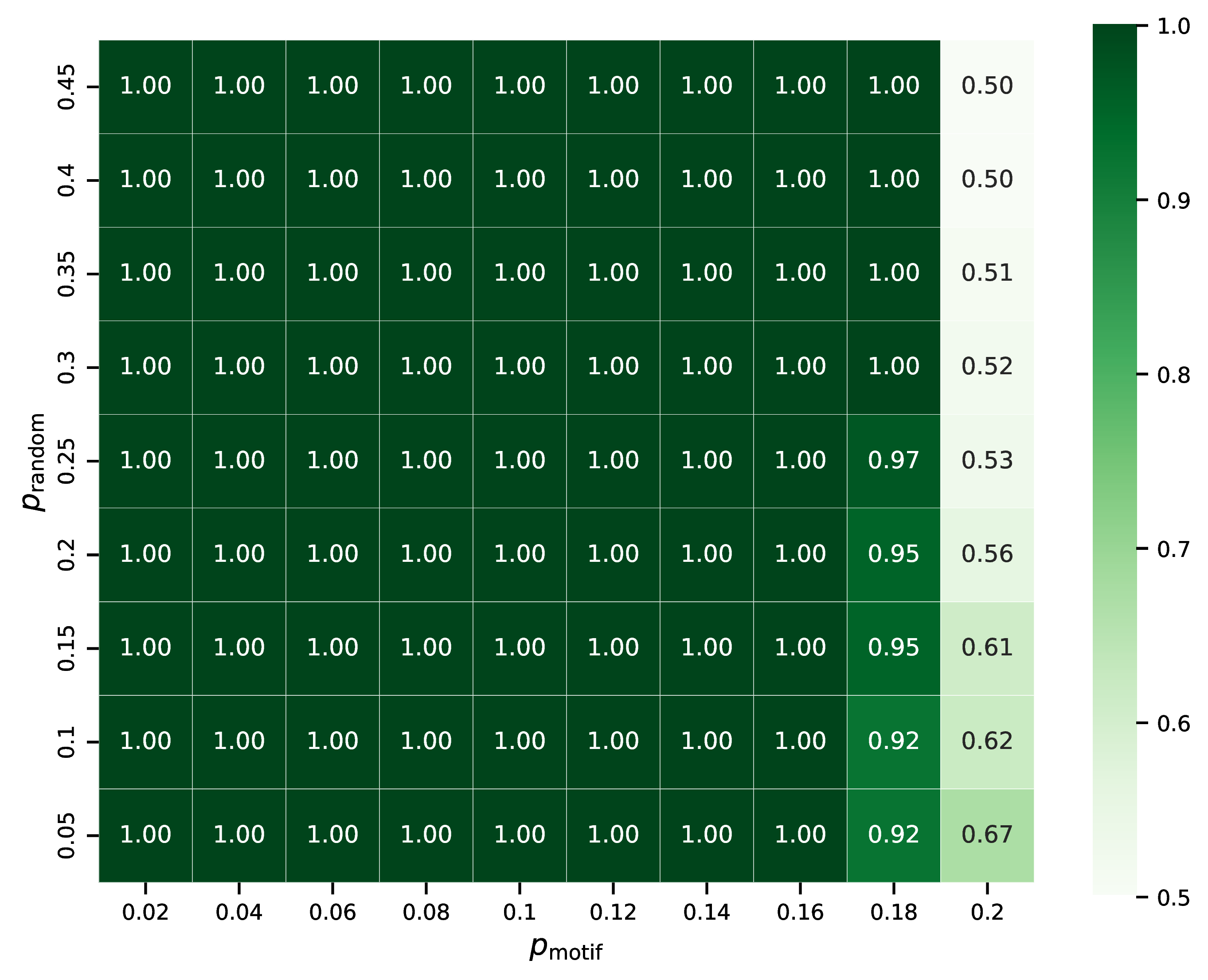}
    \caption{Test-set accuracy.}
    \label{fig:accuracy_heatmap}
    \end{subfigure}
\caption{The test-set accuracy of the smoothed classifier and the certified volume for various values of $\mathbf{p}=(p_{\text{motif}}, p_{\text{random}})$.}
\end{figure*}
\begin{figure*}
\captionsetup[subfigure]{singlelinecheck=true, labelsep=space}
\centering
    \begin{subfigure}[b]{0.79\textwidth}
    \includegraphics[height=1.90cm]{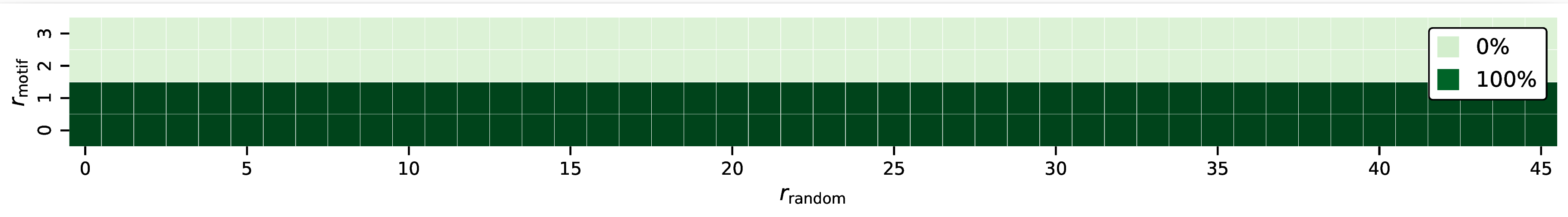}
    \caption{Anisotropic.}
    \label{fig:anis_cert}
    \end{subfigure}
    \begin{subfigure}[b]{0.20\textwidth}
    \includegraphics[height=1.90cm]{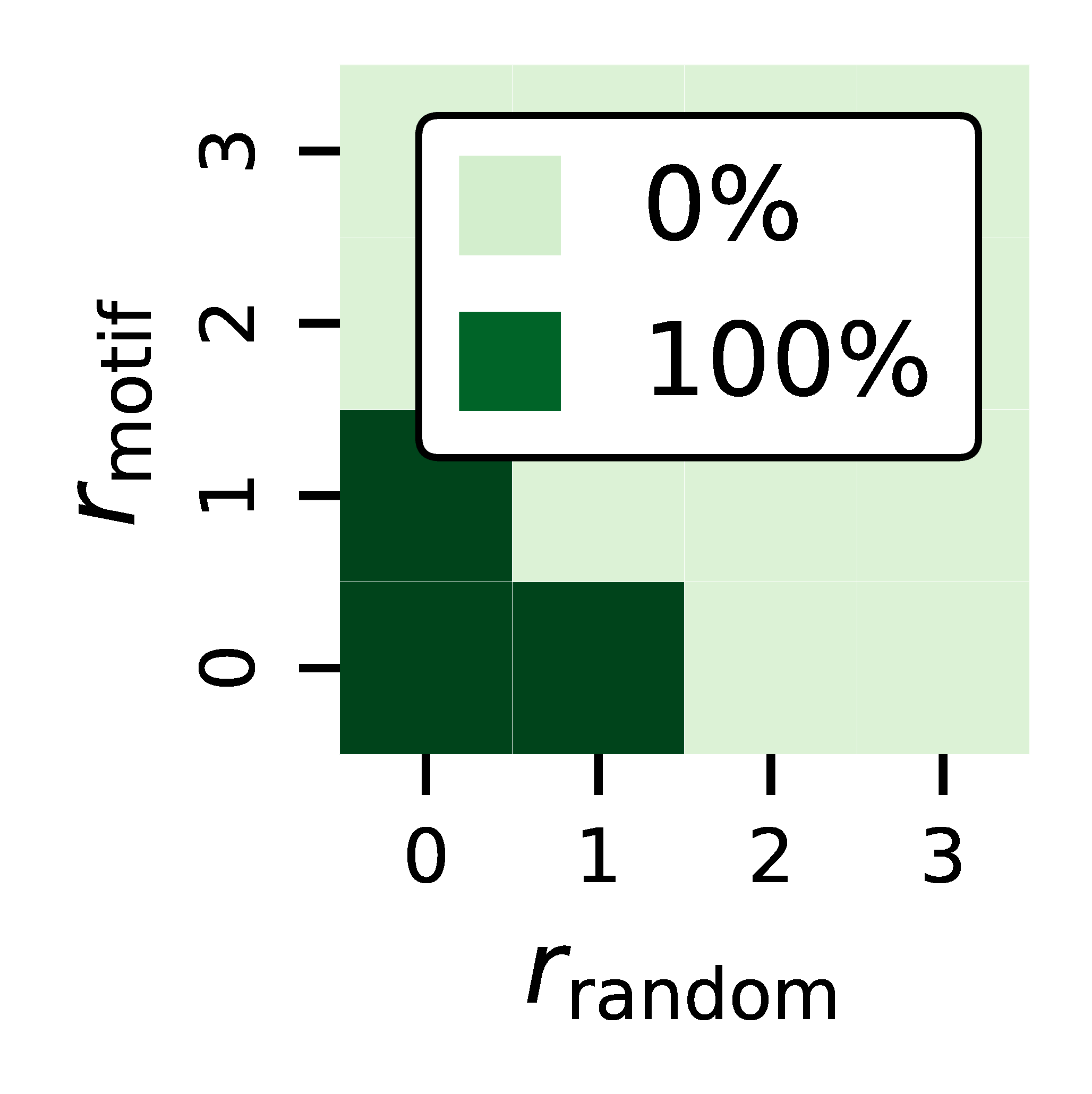}
    \caption{Isotropic.}
    \label{fig:iso_cert}
    \end{subfigure}
\caption{Comparison of an anisotropic certificates with $\mathbf{p}=(0.02, 0.45)$ and isotropic certificates for various levels of $p$. We omit some values of $p$ for the isotropic certificate for readability.} 
\end{figure*}

We first analyse how the noise vector $\mathbf{p}$ influences the certificates. We introduce a score to evaluate the noise parameters. For each $\mathbf{r}$ if we can certify strictly more than half of the test samples in this perturbation space then we add $1$ to the score. To motivate the utility of this score, consider a smoothed model with large values of noise. In the limit, a perfectly smooth classifier will be constant everywhere. In other words, it will predict the same label for all inputs and give a certified accuracy of $50\%$ for a balanced binary classification task. This classifier could be certified for arbitrary numbers of edge flips. For this reason, metrics such as total certified area averaged over samples do not necessarily tell us if a noise parameter is useful.

Fig.~\ref{fig:score_heatmap} shows our smoothed classifier has the highest score when $p_{\text{motif}}$ is small and $p_{\text{random}}$ is large. 
Fig.~\ref{fig:accuracy_heatmap} shows large values of $p_{\text{random}}$ does not effect the accuracy of the smoothed classifier, but if $p_{\text{motif}}$ becomes too large the accuracy begins to drop. These results are expected: $p_{\text{motif}}$ cannot be too large as the motif part is more important to determining the label. This motivates to fix $p_{\text{motif}}$ to be small and increase $p_{\text{random}}$, retaining high test accuracy whilst increasing the number of edge flips we can certify in $\C_2$. 

We take a closer look at the smoothed classifier given by $\mathbf{p}=(0.02, 0.45)$, one of the smoothed classifiers with the highest observed score. We are interested in a smooth model with high test set accuracy that can certify for many values of $\mathbf{r}$. Our model has $100\%$ certified accuracy. The proportion of the test set that can be certified for varying values of $\mathbf{r}$ is shown in Fig.~\ref{fig:anis_cert}. As the figure demonstrates we can certify $100\%$ of the test-set samples to $0$ or $1$ edge flips in the motif part of the graph and up to $45$ edge flips in the random part of the graph. This is the maximum number of possible node pairs in the random part, so we can certify any perturbation in this part of the graph. The smoothed classifier using isotropic noise can also achieve $100\%$ test set accuracy for all values of noise we tested. We show the certification results for when $p=0.02$, as this is the only value that allows us to certify the entire test-set for one edge flip. We plot the proportion of the test set that can be certified at using this value of isotropic noise in Fig.~\ref{fig:iso_cert}. Using larger values of noise for the isotropic certificate allows for some of the test-set to be certified at a radius of $2$, but it can no longer certify the entire test set at radius $1$. By using anisotropic noise and being specific about where edges are being certified, we can certify $46$ edge flips with $100\%$ accuracy compared to $1$ edge flip with $100\%$ accuracy.
\begin{figure*}
\centering
    \begin{subfigure}[b]{0.49\textwidth}
    \includegraphics[width=\textwidth]{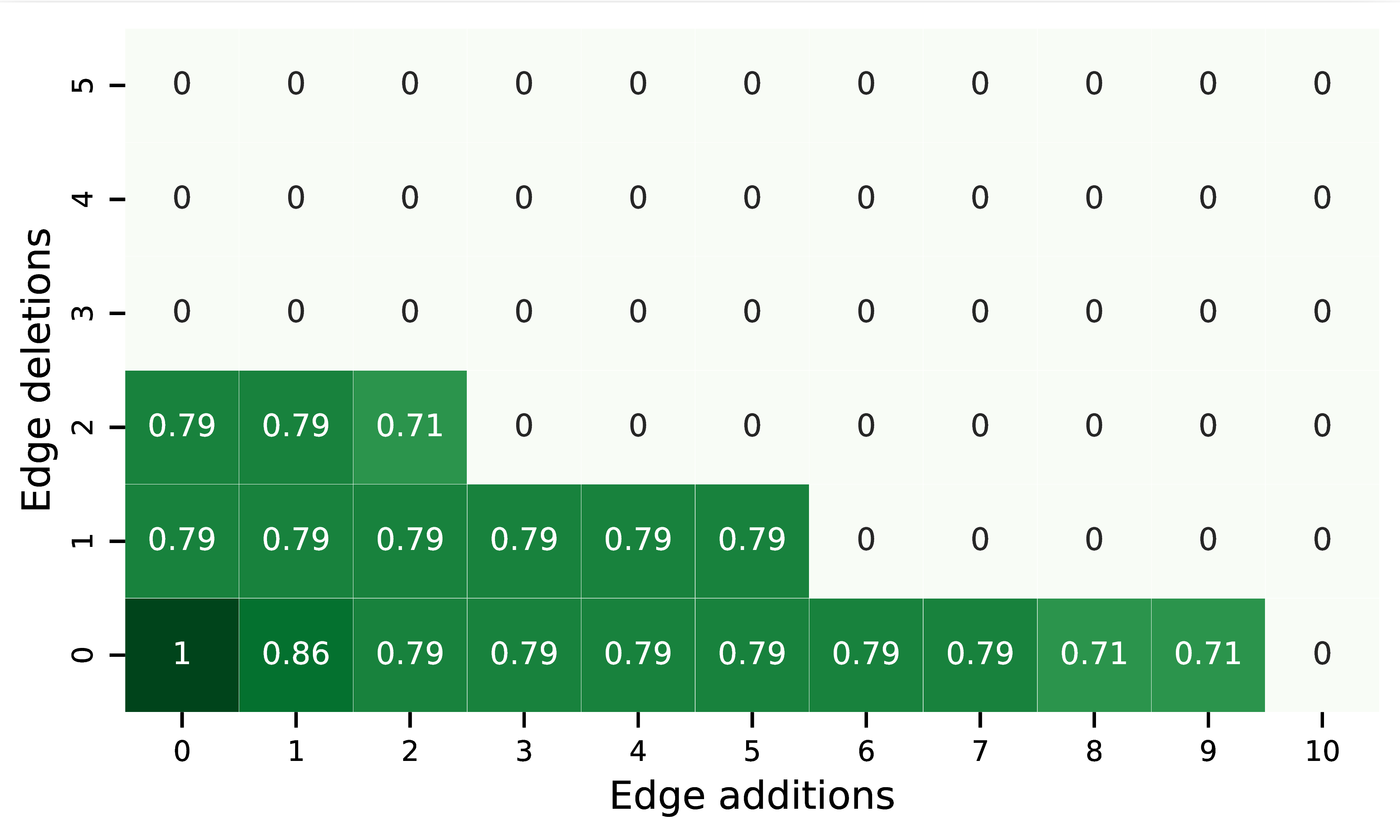}
    \caption{Anisotropic.}
    \end{subfigure}
    \begin{subfigure}[b]{0.49\textwidth}
    \includegraphics[width=\textwidth]{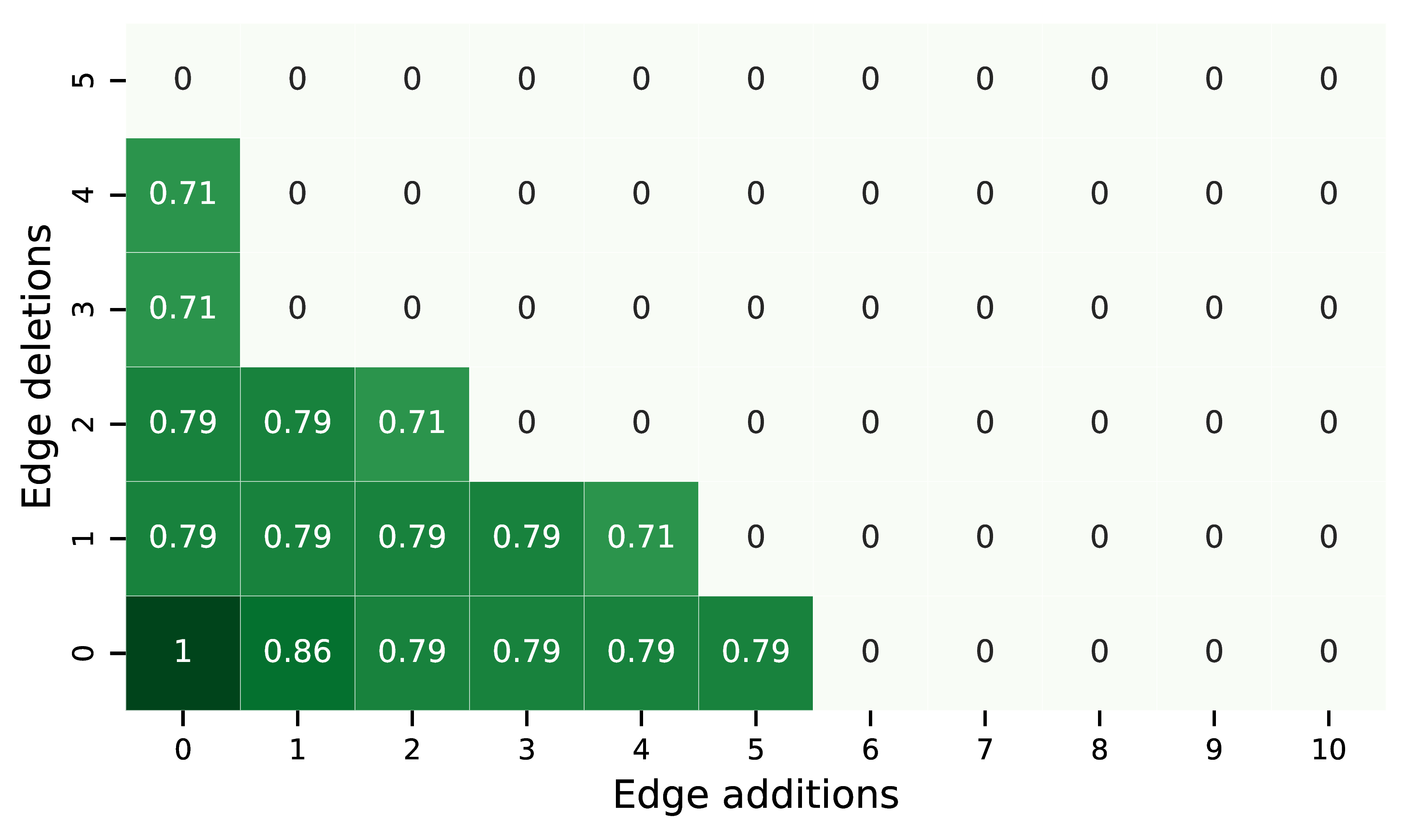}
    \caption{Sparsity-aware.}
    \end{subfigure}
\caption{A comparison between the anisotropic certificate and the sparsity-aware certificate. Each entry represents the ratio of correctly classified test-set samples that could be certified at a specified number of edge deletions and additions.}
\label{fig:comparison}
\end{figure*}

\subsection{Real-world experiment}

We also experiment using the real-world data set MUTAG  \citep{debnath1991structure}. In this data set each graph represents a molecule and the goal is to predict the molecules mutagenicity on a specific bacterium, which is encoded into a binary label. Each node has one of $7$ discrete node labels corresponding to atomic number which is one-hot encoded. The data set contains a total $188$ molecular graphs.

We train a base classifier as a graph neural network which has a single GCN layer \citep{welling2016semi} with $64$ hidden units, followed by a max pooling layer and a linear layer. We use a $80\% / 10\% / 10\%$ train/validation/test split on which to train and optimise the model hyperparameters. We train for a maximum of $500$ epochs using the AdamW optimiser \citep{DBLP:conf/iclr/LoshchilovH19} with weight decay of $10^{-3}$. The initial learning rate is $10^{-3}$ and it is decayed by $0.5$ every $50$ epochs. 

We compare our certificate to \citet{bojchevski2020efficient}, referred here as a sparsity-aware certificate, as this is the only non-isotropic certificate used for graph classification that we are aware of. We consider node pairs where there is an edge in the original graph as $\mathcal{C}_1$ and all other node pairs as $\mathcal{C}_2$. In this scenario, we can certify edge deletions and additions in a comparable way. Following the setup described in \citet{bojchevski2020efficient} we consider $p_1=0.04$ which corresponds to the probability of deleting an edge and $p_2=0.2$ which corresponds to the probability of adding an edge. We apply noise during training to make the model more robust. 

The values computed in Proposition \ref{prop:ratio} differ between the two approaches as $\P(\phi(\tilde{\x})=\z)$ is computed differently. Furthermore, the probability $\phi(\x)$ belonging to a region in the anisotropic approach is a product of Binomial distributions (Proposition \ref{prop:region}) whereas for the sparsity-aware certificate this probability follows a Poison-Binomial distribution. If the assignment of node pairs was dependent on the individual sample, this would generalise our approach further, and would also generalise the sparsity-aware certificate.

Our model has a test-set accuracy of $84\%$. In Fig.~\ref{fig:comparison} we plot the ratio of correctly predicted test points that are certified for varying numbers of edge deletions and additions. We make a few observations from these results. The first is that for values of $\mathbf{r}$ where both methods can certify test points, our method certifies the same quantity of points and in some cases more. The second is that there are two values of $\mathbf{r}$ where the sparsity-aware certificate can certify test samples but the anisotropic certificate cannot. However, there are five values of $\mathbf{r}$ where the anisotropic can certify but the sparsity-aware certificate cannot. Finally, we note that in this experiment, as well as the synthetic experiments, we find our certificates tend to be oblong, i.e. if $p_i$ is larger than we tend to certify for larger values in the $r_i$ direction. This is advantageous in the case where some node pairs are considered more important to the classification label (as demonstrated in the synthetic experiment).

\section{Conclusions}

We propose in this work, to the best of our knowledge, one of the first methods that introduces structure-aware robustness certificates in the context of classifying undirected, unweighted graphs. To achieve this, we leverage a flexible, anisotropic noise distribution in the framework of randomised smoothing and develope an efficient algorithm to compute certificates. We apply these certificates to a synthetic experiment and demonstrate a clearly improved robustness of graph classifiers that cannot be achieved with an isotropic certificates. We also validate our certificate on real-world experiments and show superior results compared to an existing approach.

Our approach requires defining a priori which edges a user believes to be more or less important to determining the graph label. Such knowledge may come from domain expertise (which we simulate in the synthetic experiment), or we may treat edge deletions and additions differently as in the sparsity-aware approach of \citet{bojchevski2020efficient}. We may also consider approaches that have been used to identify edges that may be vulnerable to attack. For example, a previous work found edges vulnerable to adversarial attack are those not captured by a low-rank approximation of the adjacency \citep{entezari2020all}. Another line of work  propose that edges where the end-point node features have low Jaccard index are potentially vulnerable \citep{wu2019adversarial}. Beyond this, one could learn the importance of the node pairs in a data-driven fashion. One such example is the recent work of \citet{sui2022causal} where the authors propose to learn the causal relations between node pairs and model outcome. We leave these directions for future work. Finally, even though we have applied our method in the context of graph classification, it can also used for any type of task based on a discrete domain such as binary image classification.


\begin{acknowledgements}
P.O and H.K acknowledge support from the EPSRC Centre for Doctoral Training in Autonomous Intelligent Machines and Systems (EP/L015897/1). X.D. acknowledges support from the Oxford-Man Institute of Quantitative Finance and the EPSRC (EP/T023333/1).
\end{acknowledgements}

%
%
%
\bibliography{references}

\appendix
\onecolumn
\section{Proofs of propositions}

\subsection{Proof of Proposition 1}

\textbf{Disjoint Unions.} Let $\z \in \R_\Q $ and $\tilde{\z} \in \R_{\Q^{'}}$ such that for some $i \in I$ we have $Q_{i} \neq Q^{'}_{i}$. If $\z = \tilde{\z}$, it implies that $||\z_{\J_{i}} - \x_{\J_{i}}|| = Q_{i}$ and $||\tilde{z}_{\J_{i}} - \x_{\J_{i}}|| = Q^{'}_{i}$ which is a contradiction.

\noindent \textbf{Partition.} $|\J_{i}| \leq R_{i}$, and $||\z_{\J_{i}} - \x_{\J_{i}}|| \leq Q_{i}$ hence $\mathcal{X} = \cup_{Q \leq R} \mathcal{R}_{Q}^{R}$.

\subsection{Proof of Proposition 2}

As the noise for each entry is independent we can decompose the probabilities as so
\begin{equation}
\frac{P(\phi(\tilde{\x}) = \z)}{P(\phi(\x) = \z)} = \prod\limits_{k \in [N]} \frac{P(\phi(\tilde{\x})_k = \z_k)}{P(\phi(\x)_k = \z_k)}. \nonumber
\end{equation}
Furthermore, as each components belongs to exactly one edge community. 
\begin{equation}
    \prod\limits_{k \in [N]} \frac{P(\phi(\tilde{\x})_k = \z_k)}{P(\phi(\x)_k = \z_k)} = \prod_{i=1}^{I} \prod\limits_{k \in \C_i} \frac{P(\phi(\tilde{\x})_k = \z_k)}{P(\phi(\x)_k = \z_k)}. \nonumber
\end{equation}
We note that for $k$ where $\tilde{\x}_k = \x_k$ this fraction is one, so we can focus on terms when $\tilde{\x}_k \not = \x_k$. In equations this can be written as 

\begin{equation}
     \prod_{i=1}^{I} \prod\limits_{k \in \C_i} \frac{P(\phi(\tilde{\x})_k = \z_k)}{P(\phi(\x)_k = \z_k)} = \prod_{i=1}^{I} \prod\limits_{k \in \J_i} \frac{P(\phi(\tilde{\x})_k = \z_k)}{P(\phi(\x)_k = \z_k)} \nonumber
\end{equation}

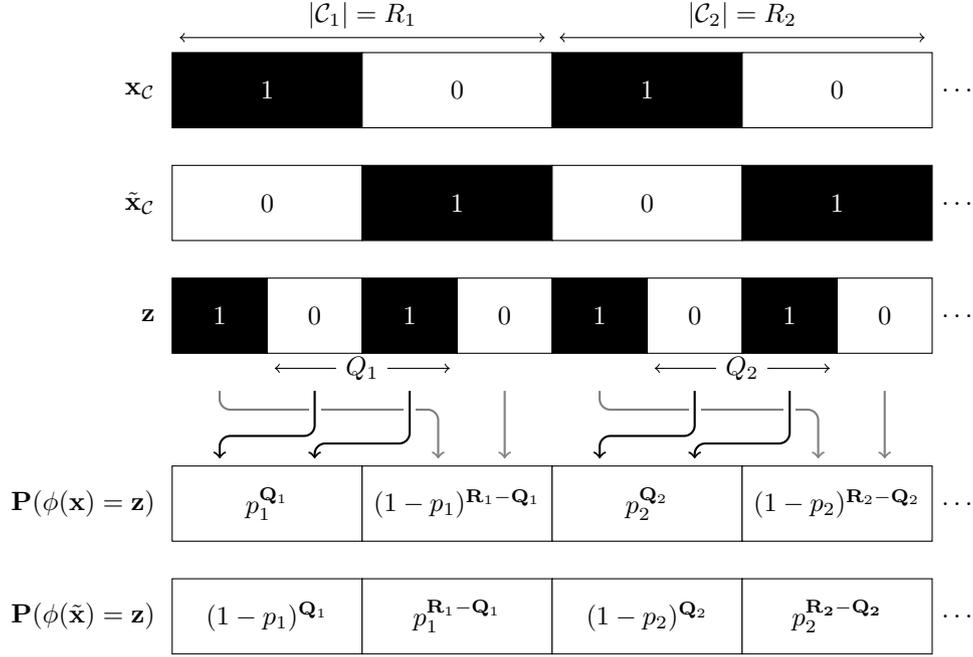
\begin{figure}[t]
\centering
\begin{tikzpicture}
\node[left] at (-0.1,0.5) {$\mathbf{x}_{\mathcal{C}}$};
\draw [fill=black] (0,0) rectangle (2.5,1) node[pos=0.5, text=white] {$1$};
\draw [fill=white] (2.5,0) rectangle (5,1) node[pos=0.5, text=black] {$0$};
\draw [fill=black] (5,0) rectangle (7.5,1) node[pos=0.5, text=white] {$1$};
\draw [fill=white] (7.5,0) rectangle (10,1) node[pos=0.5, text=black] {$0$};
\node[right] at (10,0.5) {$\ldots$};

\draw [<->] (0.1,1.2) -- (4.9,1.2) node [pos=0.5,above] {$|\mathcal{C}_1|=R_1$};
\draw [<->] (5.1,1.2) -- (9.9,1.2) node [pos=0.5,above] {$|\mathcal{C}_2|=R_2$};

\node[left] at (-0.1,-1) {$\tilde{\mathbf{x}}_{\mathcal{C}}$};
\draw [fill=white] (0,-1.5) rectangle (2.5,-0.5) node[pos=0.5, text=black] {$0$};
\draw [fill=black] (2.5,-1.5) rectangle (5,-0.5) node[pos=0.5, text=white] {$1$};
\draw [fill=white] (5,-1.5) rectangle (7.5,-0.5) node[pos=0.5, text=black] {$0$};
\draw [fill=black] (7.5,-1.5) rectangle (10,-0.5) node[pos=0.5, text=white] {$1$};
\node[right] at (10,-1) {$\ldots$};

/
/
\draw [<->] (1.35,-3.2) -- (3.65,-3.2) node [pos=0.5, fill=white] {$Q_1$};
\draw [<->] (6.35,-3.2) -- (8.65,-3.2) node [pos=0.5, fill=white] {$Q_2$};

\node[left] at (-0.1,-2.5) {$\mathbf{z}$};
\draw [fill=black] (0,-3) rectangle (1.25,-2) node[pos=0.5, text=white] {$1$};
\draw [fill=white] (1.25,-3) rectangle (2.5,-2) node[pos=0.5, text=black] {$0$};
\draw [fill=black] (2.5,-3) rectangle (3.75,-2) node[pos=0.5, text=white] {$1$};
\draw [fill=white] (3.75,-3) rectangle (5,-2) node[pos=0.5, text=black] {$0$};
\draw [fill=black] (5,-3) rectangle (6.25,-2) node[pos=0.5, text=white] {$1$};
\draw [fill=white] (6.25,-3) rectangle (7.5,-2) node[pos=0.5, text=black] {$0$};
\draw [fill=black] (7.5,-3) rectangle (8.75,-2) node[pos=0.5, text=white] {$1$};
\draw [fill=white] (8.75,-3) rectangle (10,-2) node[pos=0.5, text=black] {$0$};
\node[right] at (10,-2.5) {$\ldots$};

\draw [->, rounded corners, thick, gray] (0.625, -3.5) -- (0.625, -3.75) -- (3.5, -3.75) --  (3.5, -4.4) ;
\fill [white] (1.8,-3.8) rectangle (1.95,-3.7);
\fill [white] (3.05,-3.8) rectangle (3.2,-3.7);
\draw [->, rounded corners, thick] (1.875,-3.5) -- (1.875,-4.1) -- (0.625,-4.1) -- (0.625,-4.4) ;
\draw [->, rounded corners, thick] (3.125,-3.5) -- (3.125,-4.2) -- (1.875,-4.2) -- (1.875,-4.4) ;
\draw [->, thick, gray] (4.375, -3.5) -- (4.375, -4.4) ;

\draw [->, rounded corners, thick, gray] (5.625, -3.5) -- (5.625, -3.75) -- (8.5, -3.75) --  (8.5, -4.4) ;
\fill [white] (6.8,-3.8) rectangle (6.95,-3.7);
\fill [white] (8.05,-3.8) rectangle (8.2,-3.7);
\draw [->, rounded corners, thick] (6.875,-3.5) -- (6.875,-4.1) -- (5.625,-4.1) -- (5.625,-4.4) ;
\draw [->, rounded corners, thick] (8.125,-3.5) -- (8.125,-4.2) -- (6.875,-4.2) -- (6.875,-4.4) ;
\draw [->, thick, gray] (9.375, -3.5) -- (9.375, -4.4) ;

\node[left] at (-0.1,-5) {$\mathbf{P}(\phi(\x)=\z)$};
\draw [fill=white] (0,-5.5) rectangle (2.5,-4.5) node[pos=0.5, text=black] {$p_1^{\mathbf{Q}_1}$};
\draw [fill=white] (2.5,-5.5) rectangle (5,-4.5) node[pos=0.5, text=black] {$(1-p_1)^{\mathbf{R}_1-\mathbf{Q}_1}$};
\draw [fill=white] (5,-5.5) rectangle (7.5,-4.5) node[pos=0.5, text=black] {$p_2^{\mathbf{Q}_2}$};
\draw [fill=white] (7.5,-5.5) rectangle (10,-4.5) node[pos=0.5, text=black] {$(1-p_2)^{\mathbf{R}_2-\mathbf{Q}_2}$};
\node[right] at (10,-5) {$\ldots$};

\node[left] at (-0.1,-6.5) {$\mathbf{P}(\phi(\tilde{\x})=\z)$};
\draw [fill=white] (0,-7) rectangle (2.5,-6) node[pos=0.5, text=black] {$(1-p_1)^{\mathbf{Q}_1}$};
\draw [fill=white] (2.5,-7) rectangle (5,-6) node[pos=0.5, text=black] {$p_1^{\mathbf{R}_1-\mathbf{Q}_1}$};
\draw [fill=white] (5,-7) rectangle (7.5,-6) node[pos=0.5, text=black] {$(1-p_2)^{\mathbf{Q}_2}$};
\draw [fill=white] (7.5,-7) rectangle (10,-6) node[pos=0.5, text=black] {$p_2^{\mathbf{R_2}-\mathbf{Q_2}}$};
\node[right] at (10,-6.5) {$\ldots$};

\end{tikzpicture}
\caption{Pictorial representation of where the terms in Proposition 2 come from.}
\label{fig:prop2}
\end{figure}

We can consider what the terms are equal to when $\x_k = \z_k$ and when $\x_k \not = \z_k$ (assuming that $\x_k \not = \tilde{\x}_k$). We get 
\begin{equation}
\frac{P(\phi(\tilde{\x})_k = \z_k)}{P(\phi(\x)_k = \z_k)} = 
\begin{cases}
\frac{p_i}{1-p_i} & \text{if } \x_k = \z_k \text{ and } \x_k \not = \tilde{\x}_k \\ 
\frac{1-p_i}{p_i} & \text{if } \x_k \not = \z_k \text{ and } \x_k \not = \tilde{\x}_k \\ \nonumber
\end{cases}.
\end{equation}
In total there are $R_i$ terms in each product, of which $Q_i$ are the first case and $R_i-Q_i$ are in case two. Thus 

\begin{align}
    \prod_{i=1}^{I} \prod\limits_{k \in \J_i} \frac{P(\phi(\tilde{\x})_k = \z_k)}{P(\phi(\x)_k = \z_k)} \nonumber &= \prod_{i=1}^{I}  \left(\frac{p_i}{1-p_i}\right)^{Q_i}  \left(\frac{1-p_i}{p_i}\right)^{R_i-Q_i} \\ \nonumber
    &= \prod_{i=1}^C  \left(\frac{p_i}{1-p_i}\right)^{2Q_i-R_i} \\ \nonumber
    &= \prod_{i=1}^C  \left(\frac{1-p_i}{p_i}\right)^{R_i-2Q_i}
\end{align}
as required. We provide Fig.~\ref{fig:prop2} as a visual aid to the proof.

\subsection{Proof of Proposition 3}
 We have $\R_\Q = \set{\z \in \X : \norm{\z_{\J_{i}} - \x_{\J_{i}}}_0 = Q_{i}}$. The probability $\P(\phi(\x) \in \mathcal{R}_{\Q})$ corresponds to each set $R_{i}$ having $Q_i$ entries not being flipped or equivalently $R_i - Q_{i}$ entries being flipped. Each node pair is flipped with a probability of 
 $p_{i}$. Since all flips are independent we can express the probability as $\P(\phi(\x) \in \mathcal{R}_{\Q}) = \prod_{i = 1 }^C \Bin(R_i - Q_{i} | R_{i}, p_{i})$.

\section{Implementation}
\label{sec:implementation}

\subsection{Noise sampling}
\begin{figure}[t]
\centering
\includegraphics[width=0.8\textwidth]{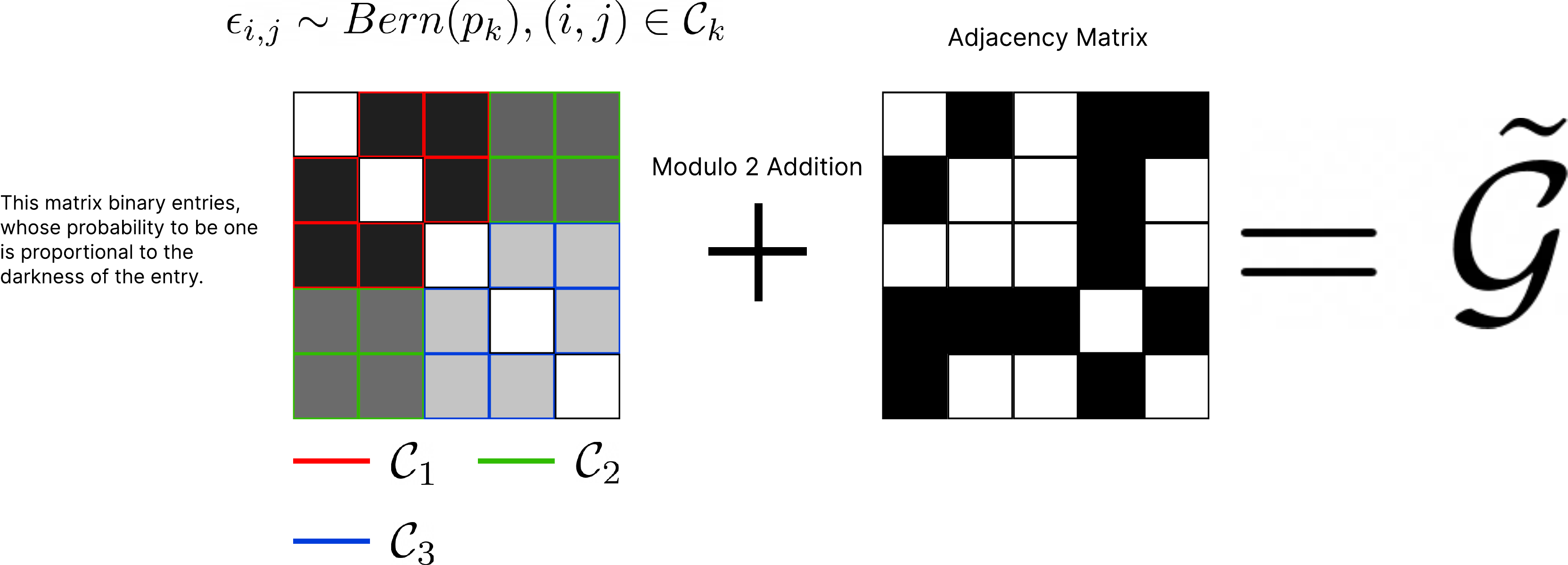}    
\caption{Illustration of the sampling procedure of the anisotropic noise distribution.}
\label{fig:sampling_anisotropic}
\end{figure}

In order to sample from the anisotropic noise defined in Eq.~(\ref{eq:smoothclassifier}), we propose an illustration in Fig.~\ref{fig:sampling_anisotropic}. Given disjoint regions of node pairs $\mathcal{C}_i$, new graphs are sampled by adding independent Bernoulli samples with parameters given by the regions to the appropriate part of the graph.
\subsection{Estimations of probabilities}

The quantities $p_{y}(\x)$ cannot be computed in closed form for general $f$. Hence, we resolve to lower bound $p_A$ and upper bound $p_{y}(\x), y \neq c_A$ via sampling. To achieve this, we use the Clopper-Pearson interval. \cite{cai2005one}.

\subsection{Symmetries certification}
Solving the optimisation problem defined in Eq.~(\ref{opti2}) is difficult as certificates have to be computed for every $\tilde \x$ in the ball around $\x$: $\mathcal{B}_{r}(\x)$. However, in practice, $\Phi_{\x, \tilde{\x}} (p_A,c_A)$ displays some symmetries depending on the noise distribution $\phi(\x)$. 

In the case of isotropic noise, the regions $\mathcal{H}_{k}$ and values $\eta_{k}$ only depends on $\norm{\x - \tilde \x}_{0}$. This implies $\Phi_{\x, \tilde{\x}} (p_A,c_A) = \Phi_{\x, \tilde{\x}^{'}} (p_A,c_A)$ for all $\tilde{\x}, \tilde{\x}^{'} \in \mathcal{S}_{r}(\x)$ which reduce the search on every spheres.

In the case of anisotropic noise, the regions $\mathcal{H}_{k}$ and values $\eta_{k}$ only depends on $\norm{\x_{\C_{i}} - \tilde{\x}_{\C_{i}}}_{0}$. This implies $\Phi_{\x, \tilde{\x}} (p_A,c_A) = \Phi_{\x, \tilde{\x}^{'}} (p_A,c_A)$ for all $\tilde{\x}, \tilde{\x}^{'} \in \mathcal{S}_{\mathbf{R}}(\x)$.
\section{Algorithm}
The full algorithm of our method is given in Alg. \ref{alg:main_alg} and its complexity is analysed below. We distinguish two use cases. We denote the first ``certification'', which is to certify a given radius $\bm R$, i.e verifying Eq.~(\ref{eq:margin_all}). The second, which we call ``optimal radius'', corresponds to solving the optimisation problem of Eq.~(\ref{opti2}).
\begin{algorithm}[ht]
\begin{footnotesize}
\caption{Structure aware randomised smoothing}
\begin{algorithmic}[1] \label{alg:main_alg}
\STATE \textbf{inputs}: Graph to certify $\x$, noise perturbation $\bm{\epsilon}$, anisotropic structure $(\mathcal{C}_i)_{i \in I} \subset [n]^{2}$, graph classification model $m: \mathcal{X} \rightarrow \mathcal{Y}$, number of samples $N$ and upper bounds on certificate radii $(\mathbf{R}_{max, i})_{i \in I}$.
\STATE \textbf{initialise}: Train model $m$ on classification data $\mathcal{D}$ or load model parameters.
\STATE \textbf{voting}
\begin{ALC@g}
\FOR{$i = 1, ..., N$}
\STATE{Sample random graph $\Tilde{\x}_i \sim \x \oplus \bm{\epsilon}$ }
\STATE{Compute model prediction $y_i \in \mathcal{Y}$}
\ENDFOR
\STATE Compute distribution label frequency from $(y_i)_{i \in [N]}$, denoted $(p_y, y)_{y \in \mathcal{Y}}$, and identify the most frequent and runner-up (second most frequent) class $(p_{A}, c_{A})$ and $(p_{B}, c_{B})$
\end{ALC@g}
\STATE \textbf{certification}
\begin{ALC@g}
\FOR{$\mathbf{R} \in \prod\limits_{i} [|\mathbf{R}_{max, i}|] $}
\STATE{Compute $\eta_{\mathbf{Q}}^{\mathcal{R}}$ according to the formula (13) and sort them.}
\STATE{Compute $\mathbb{P}(\phi(\x)) \in \mathcal{R}_{\mathbf{Q}}$ with formula (14).}
\STATE{Solve the linear programs described in Eq.~(\ref{eq:soln_upper}) and Eq.~(\ref{eq:soln_lower}) greedily}
\STATE{Verify $\underline{\rho_{\x, \tilde{\x}}}(p_A, c_A) - \overline{\rho_{\x, \tilde{\x}}}(p_B, c_B) > 0$}
\ENDFOR
\end{ALC@g}
\STATE \textbf{return} Grid of certification for $\mathbf{R} \in \prod\limits_{i} [|\mathbf{R}_{max, i}|]$
\end{algorithmic} 
\end{footnotesize}
\end{algorithm} 
\subsection{Algorithmic Complexity: Certification}

Let $\x$ be a graph, $N$ the number of samples to perform the Clopper-Pearson statistical test, $n$ the number of nodes in the graph, and $\mathbf{R} \in \prod\limits_{i} [|\mathcal{C}_i|]$ a given radius to certify and $C$ the number of regions.
The certification algorithm proceeds as follow:
\begin{enumerate}
    \item Sample $N$ graphs from the noise distribution with complexity $\mathcal{O}(Nn^{2})$, this step is very easily parallelisable.
    \item Forward the $N$ sampled graphs through the model. Given a model forward complexity of $\mathcal{O}(m(n))$ (we omit potential depency on node or edge feature dimension), the total complexity is $\mathcal{O}(Nm(n))$, this step is very easily parallelisable.
    \item From estimates $(p_A, p_B)$ and noise distribution $\mathbf{\epsilon}$ find optimal radius $\mathbf{R}$  and $T_{\mathbf{R}} = \prod\limits_{i} (R_{i} + 1)$:
    \begin{enumerate}
        \item Compute the vectors $\eta_{\mathbf{Q}}^{\mathcal{R}}$ and sort them, with respective complexity $\mathcal{O}(CT_{\mathbf{R}})$ and $\mathcal{O}(T_{\mathbf{R}} \log(T_{\mathbf{R}}))$.
        \item Solve the linear programs of Eq.~(\ref{eq:soln_upper}) and Eq.~(\ref{eq:soln_lower}) and verify $\underline{\rho_{\x, \tilde{\x}}}(p_A, c_A) - \overline{\rho_{\x, \tilde{\x}}}(p_B, c_B) > 0$, with complexity $\mathcal{O}(T_{\mathbf{R}})$.
    \end{enumerate}
    The total complexity becomes $\mathcal{O}(Nn^{2} + Nm(n) + CT_{\mathbf{R}} + T_{\mathbf{R}}\log(T_{\mathbf{R}}))$. 
\end{enumerate}
Regarding the model complexity, some example complexity are the following:
\begin{enumerate}
    \item Graph Neural Network: the complexity is quadratic in the number of nodes due to matrix multiplication: $m(n) = \mathcal{O}(n^{2})$
    \item Label kernel: the complexity is linear in the number of edges $\mathcal{O}(\mathcal{E}) = \mathcal{O}(n^{2})$ 
\end{enumerate}

\subsection{Algorithmic Complexity: Optimal radius}
Let $\x$ be a graph, $N$ the number of samples to perform the Clopper-Pearson statistical test, $n$ the number of nodes in the graph.
The algorithm to find the optimal radius proceeds as follow:
\begin{enumerate}
    \item Sample $N$ graphs from the noise distribution with complexity $\mathcal{O}(Nn^{2})$, this step is very easily parallelisable.
    \item Forward the $N$ sampled graphs through the model. Given a model forward complexity of $\mathcal{O}(m(n))$ (we omit potential depency on node or edge feature dimension), the total complexity is $\mathcal{O}(Nm(n))$, this step is very easily parallelisable.
    \item From estimates $(p_A, p_B)$ and noise distribution $\mathbf{\epsilon}$ find optimal radius $\mathbf{R}$. Select a vector, $\mathbf{R} \in \prod\limits_{i} [|\mathcal{C}_i|]$, let $T_{\mathbf{R}} = \prod\limits_{i} (R_i + 1)$ and $T = \prod\limits_{i} (R_{i, max} + 1)$:
    \begin{enumerate}
        \item Compute the vectors $\eta_{\mathbf{Q}}^{\mathcal{R}}$ and sort them, with respective complexity $\mathcal{O}(CT_{\mathbf{R}})$ and $\mathcal{O}(T_{\mathbf{R}} \log(T_{\mathbf{R}}))$.
        \item Solve the linear programs of Eq.~(\ref{eq:soln_upper}) and Eq.~(\ref{eq:soln_lower}) and verify $\underline{\rho_{\x, \tilde{\x}}}(p_A, c_A) - \overline{\rho_{\x, \tilde{\x}}}(p_B, c_B) > 0$, with complexity $\mathcal{O}(T)$
    \end{enumerate} We output the pareto front $\mathbf{R}$ according to the partial ordering $\mathbf{R} \preceq \mathbf{R}' \iff \forall i, \mathbf{R}_i \leq \mathbf{R}'_{i}$ 
\end{enumerate}

The total naive complexity is $\mathcal{O}(Nn^{2} + Nm(n) + CT^{2} + T^{2}\log(T))$. However, we want to point out there are multiple places the complexity could drastically improve.
\begin{enumerate}
    \item First, the last point is problem agnostic, meaning that, given the estimates $(p_A, p_B)$ (first and second highest label probabilities) and the noise distribution $\mathbf{\epsilon}$, the corresponding optimal radii $\mathbf{R}$ can be computed. Given specific scenario, this opens the possibility to precompute tables $\mathbf{R}(p_A, p_B, \mathbf{\epsilon})$. This can be used to directly output $\mathbf{R}$ or use it to find the optimal $\mathbf{R}$ quicker.
    \item Second, the linear program described in equation (\ref{eq:soln_upper}) and (\ref{eq:soln_lower}) can be efficiently solved greedily. Given we know the closed-form formula for $\mu_k$, making the ordering explicitly dependant on $\mathbf{Q}$, one can compute them only when necessary.
    \item Finally, the partial ordering defined previously is, in practice indicative of the robustness certification, i.e. if we cannot certify a certain radius, a larger radius won't be certified either. Although we don't propose a formal proof of this property, it holds true in practice, as one can see on the experiment results, and could be exploited for more efficient search, similar to a multidimensional binary search. 
\end{enumerate}

\section{Additional results}

\begin{figure}[t]
\centering
\includegraphics[width=\textwidth]{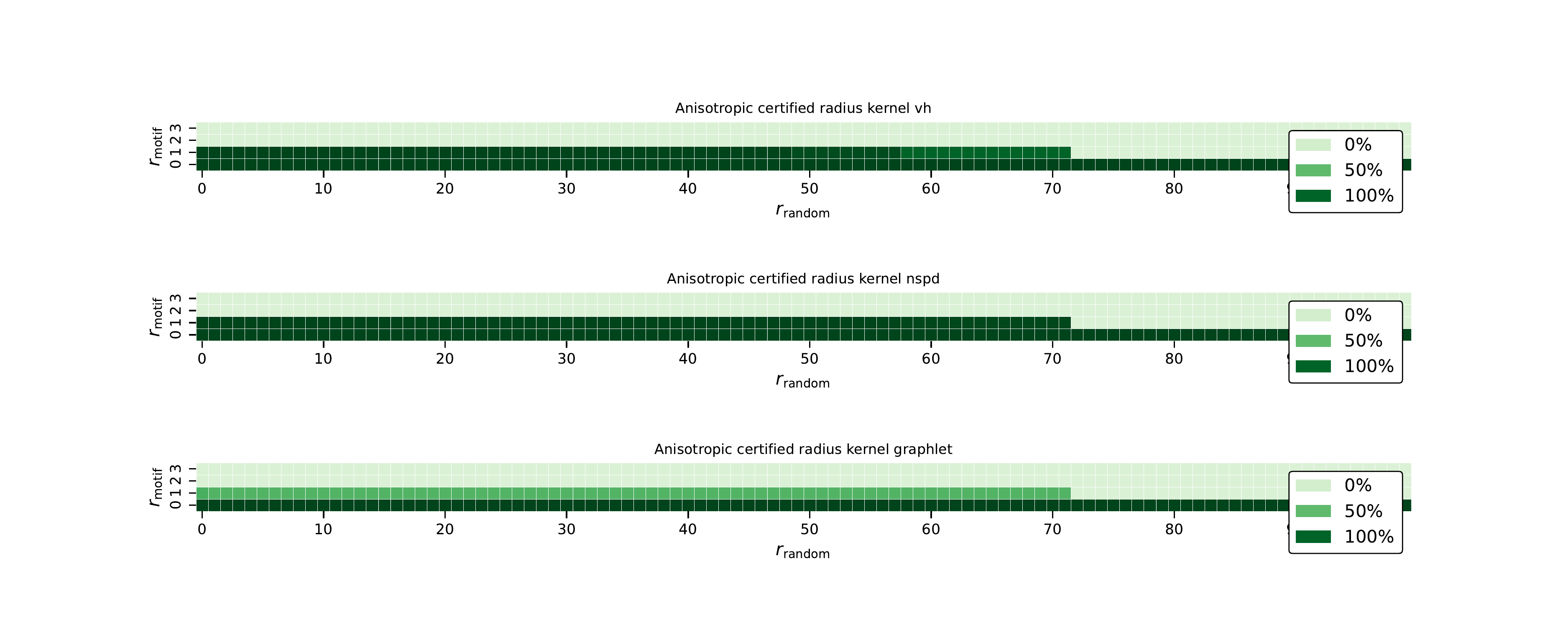}    
\caption{Influence of the underlying classifier on the anisotropic certificate radius.}
\label{fig:modelradius}
\end{figure}

\textbf{Varying the base classifier} In Fig.~\ref{fig:modelradius}, we compare our anisotropic certification performance across three kernels, the graphlet Sampling kernel \cite{shervashidze2009efficient}, the neighbourhood subgraph pairwise distance kernel \cite{costa2010fast} and vertex Histogram kernels \cite{sugiyama2015halting} for a sample size of $N=10,000$. 
In general, a model that is robust to noise will lead to  certificates with large radii.

\begin{figure}[t]
\centering
\includegraphics[width=\textwidth]{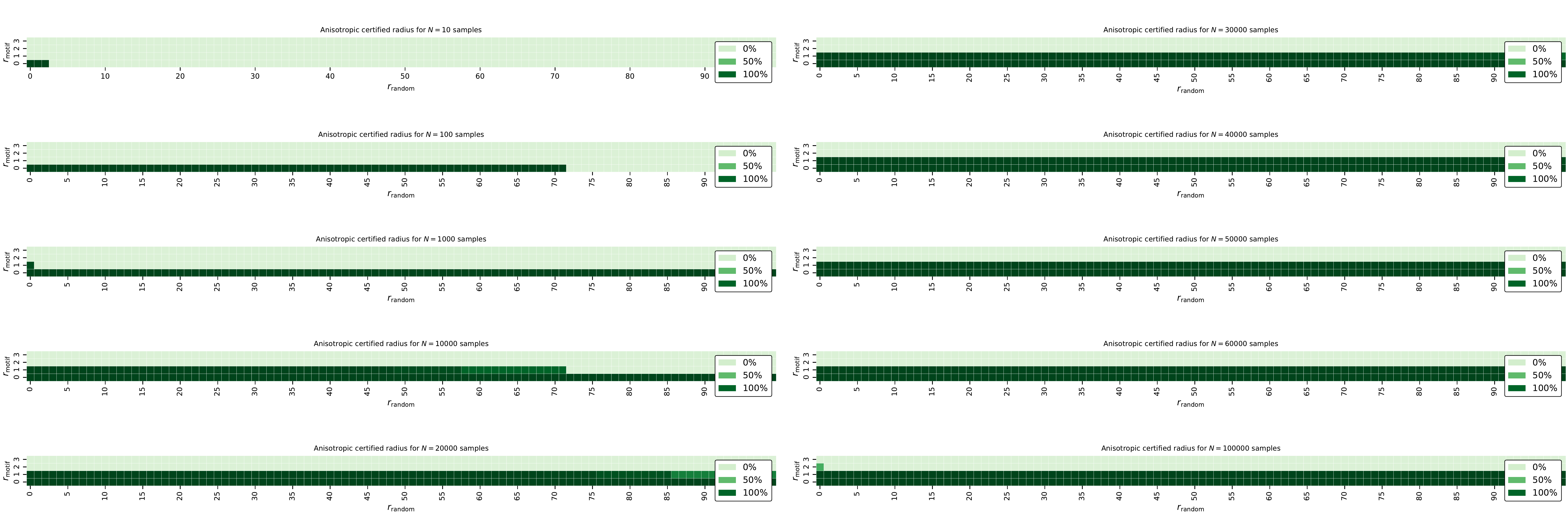}    
\caption{Influence of sample size on anisotropic certificate radius.}
\label{fig:annoisensamples}
\end{figure}

\textbf{Number of sampled perturbations} In Fig.~\ref{fig:annoisensamples}, we analysed the impact of sample size when computing the anisotropic certification radius in our synthetic experiments. 
The certificate performs poorly for a small number of samples. This is because the lower bound on $p_A$ becomes very loose.

\end{document}